%% file: paper.tex
\documentclass[]{bytedance_seed}

\usepackage[toc,page,header]{appendix}
\usepackage{calc}
\usepackage{float}

\usepackage{minitoc}
\usepackage{graphicx}
\usepackage{subcaption}
\usepackage{pifont}
\usepackage{makecell}
\usepackage{booktabs}
\usepackage{multirow}
\usepackage{float}
\usepackage{enumitem}

\usepackage{mathrsfs}
\usepackage{adjustbox}
\usepackage{multirow}
\usepackage{multirow}
\usepackage{multicol}
\usepackage{tcolorbox}
\usepackage{changepage}
\usepackage{graphicx}
\usepackage{amssymb}
\usepackage{array}
\usepackage{bm}
\usepackage{hyperref}

\usepackage{minitoc}

\newcommand{\cmark}{\ding{51}}
\newcommand{\xmark}{\ding{55}}

\newcommand{\fittowidth}[1]{%
  \sbox0{#1}%
  \ifdim\wd0>\textwidth
    \resizebox{\textwidth}{!}{\usebox0}%
  \else
    \usebox0%
  \fi
}

\title{Seedance 2.0: \\ Advancing Video Generation for World Complexity}
\author{ByteDance Seed}

\begin{document}
\maketitle


\input{sections/introduction}

\input{sections/evaluation}

\clearpage
\bibliographystyle{plainnat}
\bibliography{main}

\clearpage
\input{sections/contributions}


\end{document}

%% file: sections/introduction.tex
\section{Introduction}

Video generation models have become a core technology for modern digital content infrastructure and generative AI ecosystems, with rapid, widespread adoption across professional content production and consumer-facing creative scenarios. Through years of iterative research and engineering optimization, the ByteDance Seed team has built full-stack generative media technologies — including previous Seedance video generation series~\cite{pixeldance,seaweed,seedance1p0,seedance1p5}, Seedream image generation and editing series~\cite{seedream2,seedream3,seedream4,seededit,seededit3p0,mogao}, Seed-VL multimodal vision-language models~\cite{seed1p5,bagel,seed2p0} for cross-modal semantic understanding, and other key components~\cite{rewarddance,dancegrpo}, which are now widely integrated into our large-scale product ecosystem, supporting video generation services for billion-level daily active users.

In this work, we push the frontier of video generation technologies with a notable paradigm shift: from generating short video clips with limited controllability to robust, highly controllable video synthesis natively supporting diverse control signals. This industry-wide trend has driven the development of Seedance 2.0, our new model designed specifically to deliver enhanced generation quality with rich multi-modal controllability for large-scale creative engine platforms.


Seedance 2.0 is a new \textbf{native multi-modal audio-video generation model}, officially released in China in early February 2026. Compared with its predecessors, Seedance 1.0 and 1.5 Pro~\cite{seedance1p0, seedance1p5}, Seedance 2.0 adopts a unified, highly efficient, and large-scale architecture for multi-modal audio-video joint generation. This allows it to support four input modalities: text, image, audio, and video, by integrating one of the most comprehensive suites of multi-modal content reference and editing capabilities available in the industry to date. It delivers substantial, well-rounded improvements across all key sub-dimensions of video and audio generation. In both expert evaluations and public user tests, the model has demonstrated performance on par with the leading levels in the field.

Within its multi-modal framework, Seedance 2.0 is equipped with a full set of multi-modal reference and editing capabilities, supporting both standalone and combinatorial tasks, including subject control, motion manipulation, style transfer, special effects design and creative content generation, as well as video extension. This suite of capabilities renders the model applicable to a diverse array of creative production scenarios. The model achieves more accurate compliance with complex instructions and more stable motion performance, enabling it to accommodate more sophisticated production workflows. Coupled with its native, professional multi-shot narrative capability, vivid details of motion and facial expressions, and improved cross-frame consistency, the model reaches a competitive usability rate in real-world production settings.

Seedance 2.0 supports direct generation of audio-video content with durations ranging from 4 to 15 seconds, with native output resolutions of 480p and 720p. For multi-modal inputs as reference, its current open platform supports up to 3 video clips, 9 images, and 3 audio clips. In addition, we provide \textbf{Seedance 2.0 Fast} version, an accelerated variant of Seedance 2.0 designed to boost generation speed for low-latency scenarios. Seedance 2.0 has delivered significant improvements to its foundational generation capabilities and multi-modal generation performance, bringing an enhanced creative experience for end users. Its key model capabilities are highlighted as follows.

\begin{itemize}[leftmargin=0.5cm]

\item \textbf{Generation of Real-world Complexity.}
Seedance 2.0 achieves remarkable improvements in generation quality, particularly in human motion modeling, which delivers significantly enhanced naturalness, temporal coherence, and physical plausibility compared to previous versions. It can synthesize temporally precise, complex interaction scenes with high fidelity, while adhering to real-world motion laws throughout the generation process, thereby mitigating the artifacts common in recent video generation models. For detailed close-up shots, the generated frames exhibit highly realistic details and rigorous consistency—whether for subtle changes in light refraction or natural, fluid interactions between characters and the environment—closely matching the visual fidelity of real-world live-action footage. With robust motion stability and physics compliance, the model delivers favorable performance in multi-subject interaction and complex motion scenarios, achieving a usability rate that is clearly higher than recent commercial models.

\item \textbf{Strong multimodal capability.} 
First, Seedance 2.0 accepts a comprehensive multimodal reference input, allowing you to combine text, image, video, and audio sources. The model can accurately interpret multimodal input content and generate output referencing user-specified elements that include frame composition, cinematographic design, motion rhythm, and acoustic characteristics in accordance with user instructions. It can also directly reference text-based storyboards, enhancing the flexibility of conventional video generation workflows and expanding the creative freedom.
%
%
%
Second, Seedance 2.0 features substantially improved controllability in video generation. It delivers strong instruction-following performance, accurately generating specified content, and maintaining consistent subject identity preservation even when processing complex scripts with extensive character interactions and fine-grained action descriptions. 
Meanwhile, the model exhibits fundamental directorial and cinematographic reasoning capabilities, enabling it to autonomously plan shot sequencing and design visual presentation templates. 
In addition, Seedance 2.0 introduces new video editing capabilities, which enable targeted modifications to specified clips, characters, actions, or plot elements. It also provides video continuation functionality, which generates consecutive shots aligned with user prompts, supporting both de novo video generation and seamless extension of existing footage.

\item \textbf{High-Fidelity Audio-Video Generation.} 
Seedance 2.0 has binaural audio capability with synchronized high-fidelity immersive sound generation. It is equipped with an upgraded audio generation module integrated with binaural audio technology, which enables high-fidelity, immersive sound generation. The model supports simultaneous multi-track output of audio content including background audio, ambient sound effects, and character narration, with precise temporal alignment to the visual rhythm of generated footage. The audio content generated by the model features highly natural sound design, faithfully reproducing subtle natural ambient sounds to enhance scene immersion. Coupled with strict audio-visual temporal control, the model ensures tight synchronization between audio tracks and visual actions, providing robust support for professional-grade audio-visual content creation.

\item \textbf{Applications in Productivity Scenarios.}
It exhibits strong cross-scene adaptability, which reduces the barriers to professional content production.
In response to the diverse demands of video content production, Seedance 2.0 demonstrates high cross-scene adaptability. It delivers high-quality generation results across a wide range of use cases, including commercial advertising, cinematic and television visual effects, game animation, and commentary videos. By replacing complex visual effects production and live-action shooting workflows with AI generation, Seedance 2.0 can significantly reduce production costs and shorten the production cycle of professional audio-video content, helping creators and enterprises realize their creative concepts.

\end{itemize}

From the audio-video synchronous generation achieved by Seedance 1.5~\cite{seedance1p5} to the unified multimodal audio-video joint generation framework established by Seedance 2.0, the Seedance series has consistently been built around a unified architecture, with a core commitment to high-fidelity reconstruction of real-world complexity.

We acknowledge that Seedance 2.0 remains imperfect, with room for improvement in its generation outputs. Moving forward, we will continue to explore deep alignment between generative models and the physical world, advance accurate modeling of real-world dynamics, deepen our understanding of physical and semantic rules, and enable the technology to better serve every creator.

Safety is a core consideration in our work. 
Throughout the model iteration lifecycle, we have implemented a structured safety assessment framework and made continuous efforts to evaluate and mitigate potential risks, with the aim of supporting responsible, compliant, and ethically aligned development.

\begin{tcolorbox}[colback=gray!4!white, colframe=gray!25!white, arc=5pt, boxrule=0.4pt]
We invite readers to explore the capabilities of Seedance 2.0.

Seedance 2.0 is now accessible on Doubao\footnotemark[1], Jimeng\footnotemark[2] and Volcano Engine, under the model id: doubao-seedance-2-0-260128.

The model can be accessed at \url{https://www.volcengine.com/experience/ark?mode=vision&modelId=doubao-seedance-2-0-260128&tab=GenVideo}.

More details are available on the official page: \url{https://seed.bytedance.com/seedance2_0}.
\end{tcolorbox}

\footnotetext[1]{https://www.doubao.com/chat/create-video}
\footnotetext[2]{https://jimeng.jianying.com/ai-tool/video/generate}

%% file: sections/evaluation.tex
\section{Evaluation}

\subsection{Overview}
\begin{figure}[H]
    \centering
    \includegraphics[width=1.0\textwidth]{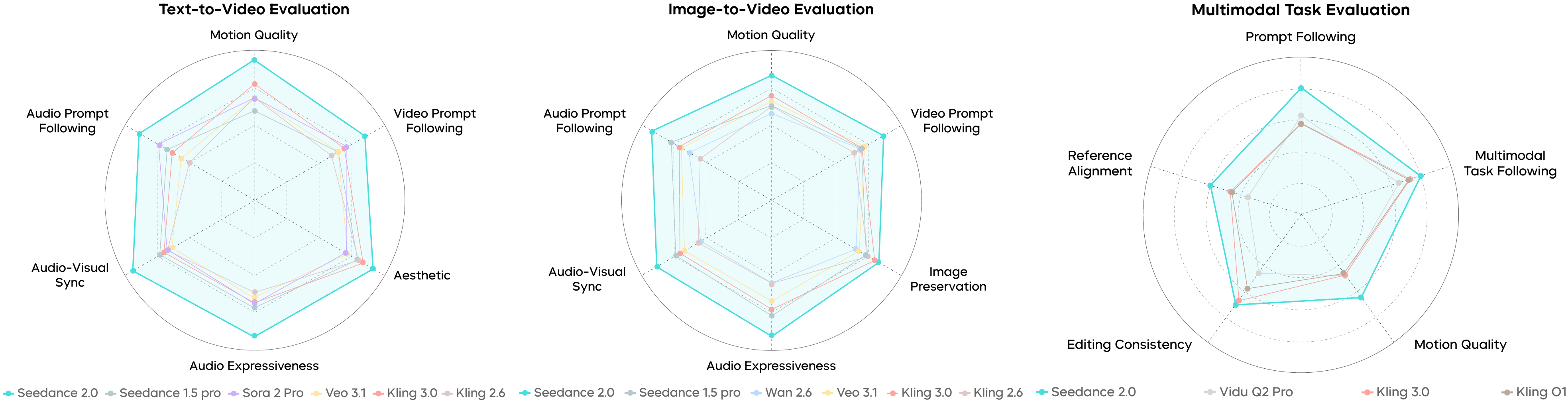}
    \caption{Overall performance comparison across T2V, I2V, and R2V tasks. Seedance 2.0 achieves comprehensive leading performance over all competing models across every evaluated dimension in all three generation tasks.}
    \label{fig:overall_radar}
\end{figure}

To objectively and comprehensively assess the overall capabilities of Seedance 2.0 in multi-modal scenarios, our team collaborated with experts from the media industry to establish a comprehensive evaluation benchmark and corresponding evaluation protocols. The benchmark covers audio-video generation, reference-based generation, and video editing scenarios. This evaluation focuses on the model’s performance across core dimensions: multi-modal reference-based generation, complex audio-video instruction following, complex motion stability, professional cinematographic language expression, audio and video expressiveness, and audio-visual integrated alignment. As shown in Figure~\ref{fig:overall_radar}, Seedance 2.0 achieves the highest scores across all evaluated dimensions in Text-to-Video (T2V), Image-to-Video (I2V), and Reference-to-Video (R2V) tasks, demonstrating comprehensive leading performance over current competing models.

\textbf{Text-to-Video and Image-to-Video Evaluation.} In video generation tasks, Seedance 2.0 delivers competitive leading performance in the industry. Marked improvements have been achieved in motion stability, instruction following capability, and visual aesthetics. The model effectively mitigates common structural inaccuracies and visual artifacts, generating smooth and nuanced complex motions. It can accurately render high-tension large-scale movements and subtle micro-expressions, while supporting professional-level combined camera movements and narrative rhythm control.
For long scripts and open-ended instructions, the model can respond appropriately and deliver reasonable generation outputs. Meanwhile, the generated videos exhibit notable cinematic aesthetics, with well-rendered object textures, lighting and composition, as well as costume, makeup and prop design.
In the audio domain, Seedance 2.0 maintains strong performance with substantial improvements in audio expressiveness. Its dual-channel audio output presents rich and nuanced layers, and can generate sound effects or melodies that align well with the context described in the prompts. Compared with the previous version, the model delivers an enhanced audio-visual integrated experience, with tighter alignment between dialogue lines, sound effects, background audio, and visual content. Meanwhile, the instruction following accuracy for Chinese dialects, traditional opera, and singing scenarios is significantly improved.

\textbf{Multi-modal Reference-Based Generation Evaluation.}
Seedance 2.0 achieves competitive leading comprehensive performance in reference-based generation tasks. The model supports a more comprehensive range of reference-based tasks, covering multiple creative scenarios including multi-modal reference-based generation, video editing, and video continuation. Meanwhile, it demonstrates advantages in the depth of understanding and response accuracy for reference content. Specifically, in video editing tasks, Seedance 2.0 delivers more complete instruction following and more photorealistic visual outputs compared with peer models.
In terms of generation consistency, the model achieves favorable performance in subject appearance and voice restoration, with particularly notable advantages in maintaining reference consistency for action logic, special effect styles, and plot narrative. Despite these strengths, there is still room for optimization in multi-subject consistency, text restoration accuracy, and the performance of complex editing tasks.

We evaluate Seedance 2.0 on three generation tasks---T2V (Text-to-Video), I2V (Image-to-Video), and R2V (Reference-to-Video)---against current commercial video generation models. Seedance 2.0 ranks first across all video and audio dimensions on every task.

On video, the main advances are: (1) Stability---fewer deformations and structural issues; complex actions are fluid; multilingual text generation and preservation are relatively strong. (2) Vividness---sports, combat, and other high-amplitude actions carry strong momentum; facial expressions and gaze are emotionally engaging; the model produces professional-level camera movements, dynamic editing, and narrative pacing. (3) Instruction following---long-script prompts are executed with reasonable precision, multi-shot and multi-angle instructions are followed accurately, and open-ended prompts receive appropriate creative interpretation; multiple art styles are supported; in I2V, special art styles from the reference image are preserved while adapting subject motion to match. (4) Visual realism---object materials look authentic, lighting, composition, and character texture improve notably, and costume and prop design is polished.

On audio: (1) Expressiveness---audio is detailed and layered, with dual-channel support; melodies and tones match the prompt context, and the audio dimension adds to the emotional impact of the video. (2) Audio-visual sync---lip movements match the visual, dialogue, sound effects, and background audio align well, and beat-matching between audio and video is strong. (3) Audio instruction following---Chinese dialects (Sichuan, Northeastern, Cantonese), opera, and singing improve markedly over the previous version; singing, rap, and instrumental performance are consistent, with melodies adapted to the prompt context.

Areas for improvement remain: minor deformation artifacts, motion plausibility in edge cases, high-frequency visual noise, audio distortion and noise, and lip-sync errors in multi-speaker scenes. The following subsections present detailed results for each task.

\subsection{Evaluation Framework}

To prepare the model for production deployment, we upgraded our evaluation framework to SeedVideoBench 2.0. The new version adds multimodal generation, narrative quality, and multilingual coverage to the evaluation scope, and refines how audio expressiveness is assessed. We also brought in expert evaluators from advertising and game production to provide subjective ratings, with a focus on narrative and aesthetic quality.

\subsubsection{SeedVideoBench 2.0}

Two main changes define the new framework. First, a multimodal task evaluation system that formally defines multimodal task following and generation consistency, while also covering baseline generation quality (prompt following, motion quality) in multimodal settings. Second, we split evaluation into objective and subjective tracks. Objective metrics like motion stability---use automated pipelines. Subjective metrics like aesthetics---go through blind expert review. Separately, we ran a realism study: evaluators tried to tell Seedance 2.0 outputs apart from real video clips. The results fed back into our aesthetic tuning process.

The multimodal task evaluation module and the narrative assessment module saw the largest changes from the previous version. \textbf{Multimodal Task Following} measures instruction-following accuracy across reference, editing, and extension scenarios, broken into dozens of fine-grained task types (subject identity, motion, style, etc.). Most models have limited multimodal coverage, which forces users to probe capability boundaries through trial and error; these metrics make the boundaries explicit. Specifically, the evaluation covers four task groups:
\begin{itemize}[leftmargin=0.5cm]
  \item \textit{Reference tasks:} subject, motion, visual-effects, and style reference generation.
  \item \textit{Editing tasks:} subject, style, scene, and audio content editing.
  \item \textit{Extension tasks:} plot continuation and seamless extension, both forward and backward on the timeline.
  \item \textit{Combination tasks:} paired evaluations that match real workflows---e.g., swapping a video subject with a reference image (reference + editing combined).
\end{itemize}
\noindent \textbf{Consistency} captures how closely generated content matches the reference input (reference alignment) and how well non-edited regions survive during editing (editing consistency). We built specialized datasets covering subject, motion, scene, style, and audio, with sample distributions tuned to minimize variance at small evaluation budgets.

On the video metrics side, SeedVideoBench 1.5 already tracked vividness; version 2.0 adds finer \textbf{narrative quality} metrics alongside the existing vividness and aesthetics dimensions. Unlike motion quality, which can be measured more objectively, narrative quality is inherently subjective---it asks whether the overall narrative reads as coherent, whether character performances and visual effects carry emotional weight, and whether the aesthetic choices fit the content. We evaluate it along three sub-dimensions:

\begin{itemize}[leftmargin=0.5cm]
  \item \textit{Cinematographic language:} does the camerawork support the story? We assess shot logic and expressiveness, looking for problems like redundant coverage, axis-crossing (180-degree rule violations), mismatched shot sizes, and uneven pacing.
  \item \textit{Plot design:} can the model take a vague or brief prompt and produce something both coherent and engaging?
  \item \textit{Stylistic aesthetics:} do the visuals have a considered look? This covers lighting, framing, composition, and color grading, plus whether characters, costumes, props, and sets hold together aesthetically.
\end{itemize}

Unless otherwise noted, all evaluation results reported in the following sections are obtained using SeedVideoBench 2.0.

\subsubsection{Arena.AI Results}

\label{sec:ai_arena}
\begin{figure}[H]
    \centering
    \begin{subfigure}{\textwidth}
        \centering
        \includegraphics[width=1.0\textwidth]{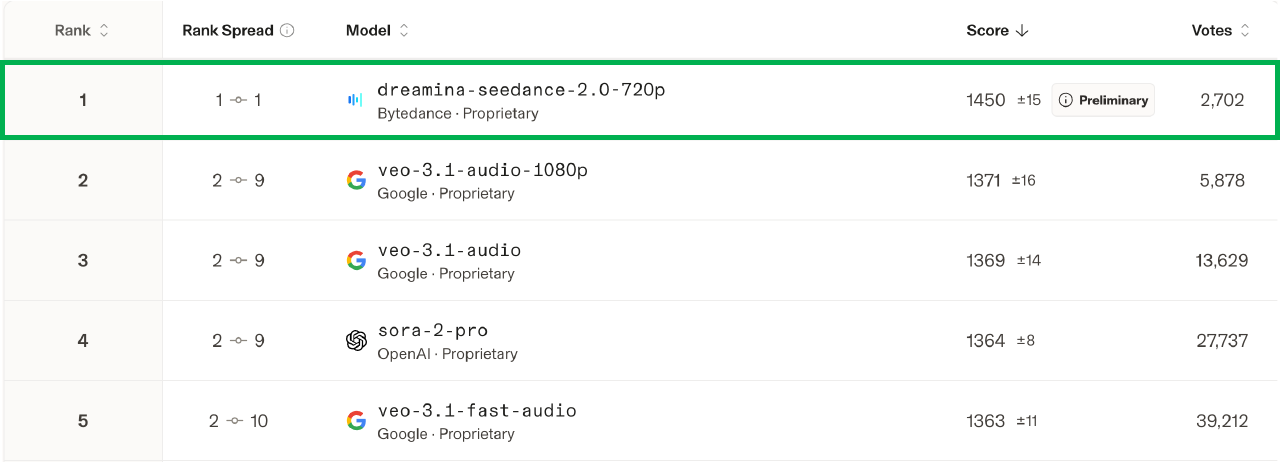}
        \caption{Text-to-Video generation leaderboard.}
        \label{fig:t2va_aiarena}
    \end{subfigure}
    \vspace{1em}
    \begin{subfigure}{\textwidth}
        \centering
        \includegraphics[width=1.0\textwidth]{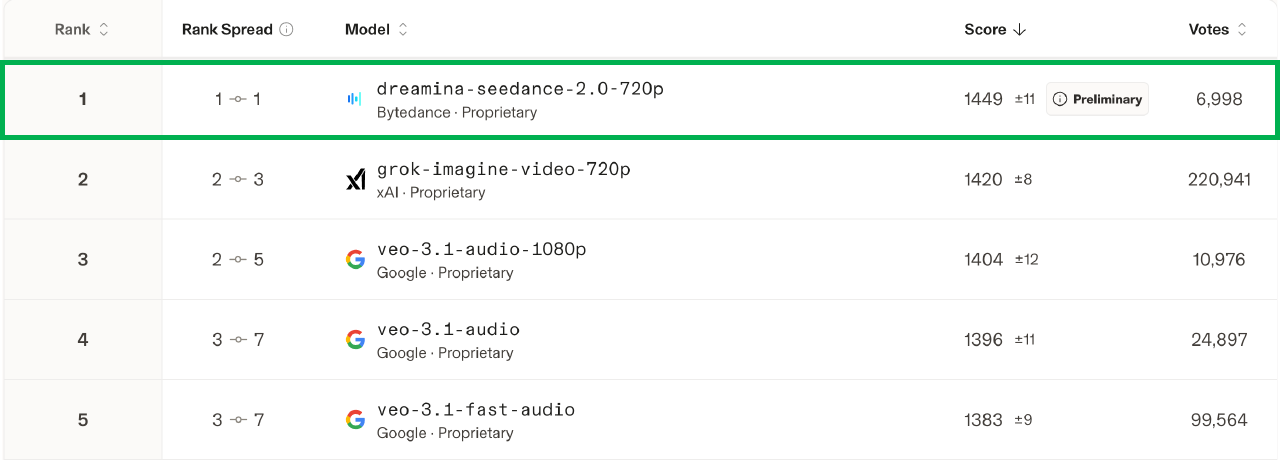}
        \caption{Image-to-Video generation leaderboard.}
        \label{fig:i2va_aiarena}
    \end{subfigure}
    \caption{Leaderboards on Arena.AI (Accessed: April 8, 2026, Eastern Time).}
    \label{fig:aiarena_leaderboards_all}
\end{figure}

Arena (formerly LMArena)~\cite{aiarena}, created by researchers from UC Berkeley, is a community-powered platform that evaluates AI models through real-world user preferences. For video generation, users are presented with outputs from two anonymous models side-by-side and vote for the one they prefer, producing an Elo-style leaderboard that reflects genuine human judgment at scale. Unlike automated benchmarks that rely on metrics such as FVD or CLIPScore, Arena captures holistic human preferences encompassing visual quality, motion realism, temporal coherence, and prompt adherence in a single unified ranking.

As shown in Figure~\ref{fig:aiarena_leaderboards_all}, our Dreamina Seedance 2.0 720p ranks \#1 on both the Text-to-Video and Image-to-Video leaderboards, with Elo scores of 1450 ($\pm$15) and 1449 ($\pm$11) respectively. On T2V (Figure~\ref{fig:t2va_aiarena}), it leads the second-place veo-3.1-audio-1080p by 79 points; on I2V (Figure~\ref{fig:i2va_aiarena}), it leads grok-imagine-video-720p by 29 points. Notably, the model achieves this at 720p resolution, outperforming competitors that operate at 1080p, which suggests that our improvements in motion dynamics and visual coherence are more perceptually significant than resolution alone. The Rank Spread of 1$\leftrightarrow$1 on both leaderboards indicates consistently top-ranked performance across evaluation dimensions. These results complement our SeedVideoBench2.0 findings, demonstrating that the gains observed in objective metrics translate directly into stronger human preference.

\subsection{Text-to-Video Evaluation on SeedVideoBench 2.0}

\subsubsection{Overall Results}

\vspace{-0.4cm}
\input{tables/t2v/t2v_overall}
\vspace{-0.7cm}
\input{tables/t2v/t2v_usability_rate}
\vspace{-0.1cm}
Table~\ref{tab:t2v_mos} summarizes the T2V results. Seedance 2.0 ranks first on all six dimensions, the only model above 3.4 on every dimension and improving over Seedance 1.5~\cite{seedance1p5} by an average of 0.86 points, with the largest gain on motion quality (+1.36). On both motion quality and audio-visual sync, Seedance 2.0 reaches 3.75, at least 0.65 points ahead of the runner-up. The audio dimensions are where competitors struggle most---most stay below 2.9---while Seedance 2.0 exceeds 3.5 on all three. Among competitors, Kling 3.0~\cite{kling3.0} is the most balanced overall, Sora 2 Pro~\cite{sora2} stands out on prompt following, and Veo 3.1~\cite{veo3p1} is weaker on audio. 

The usability breakdown in Table~\ref{tab:t2v_rates} sharpens this picture. Seedance 2.0 is the only model with usability above 83\% on all dimensions, reaching 97.55\% on motion quality. More tellingly, Seedance 2.0 exceeds 51\% satisfaction on every dimension---the majority of its outputs score 4 or above---while no competitor exceeds 44\% on any single dimension. The gap is most pronounced on audio: audio quality satisfaction is 62.05\% vs.\ below 10\% for all competitors, and audio-visual sync reaches 68.30\% vs.\ a competitor high of 25.45\%. At the delight level (score of 5), only Seedance 2.0 produces any delight-rated audio quality outputs (6.70\%), and its audio prompt following delight rate of 26.92\% far exceeds the next-best Sora 2 Pro (11.68\%).

\begin{figure}[H]
    \centering
    \begin{subfigure}[b]{0.49\textwidth}
        \includegraphics[width=\textwidth]{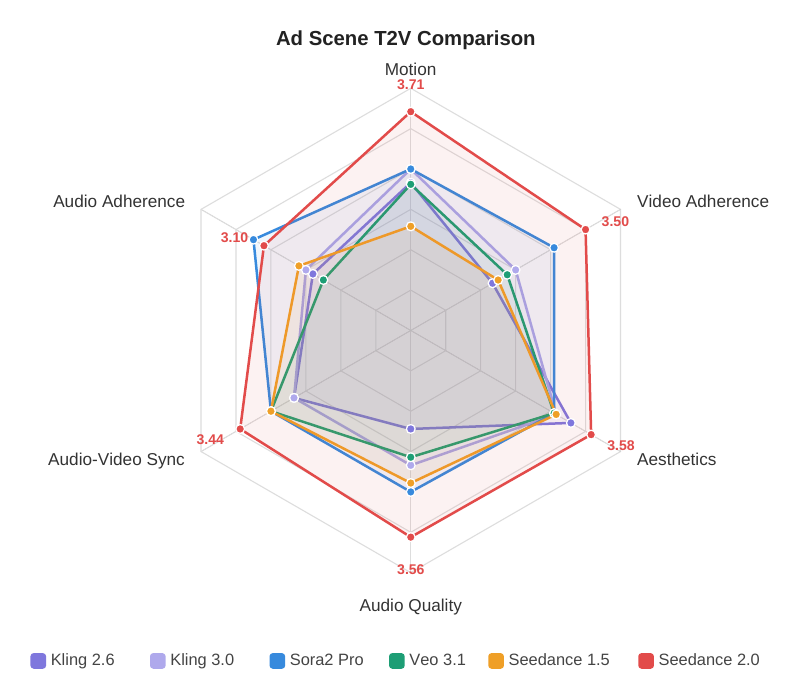}
        \caption{Ad Scene}
    \end{subfigure}
    \hfill
    \begin{subfigure}[b]{0.49\textwidth}
        \includegraphics[width=\textwidth]{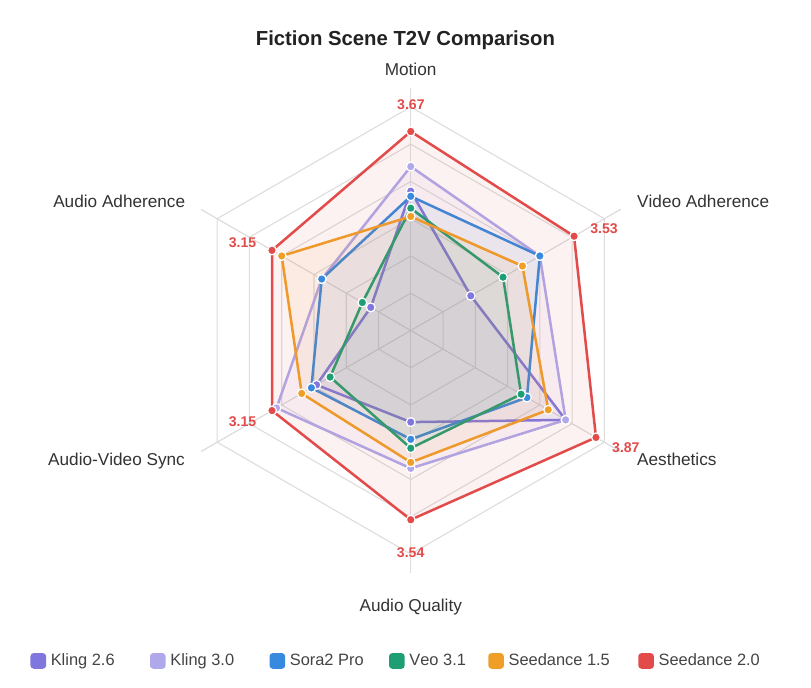}
        \caption{Fiction Scene}
    \end{subfigure}

    \vspace{0.1em}

    \begin{subfigure}[b]{0.49\textwidth}
        \includegraphics[width=\textwidth]{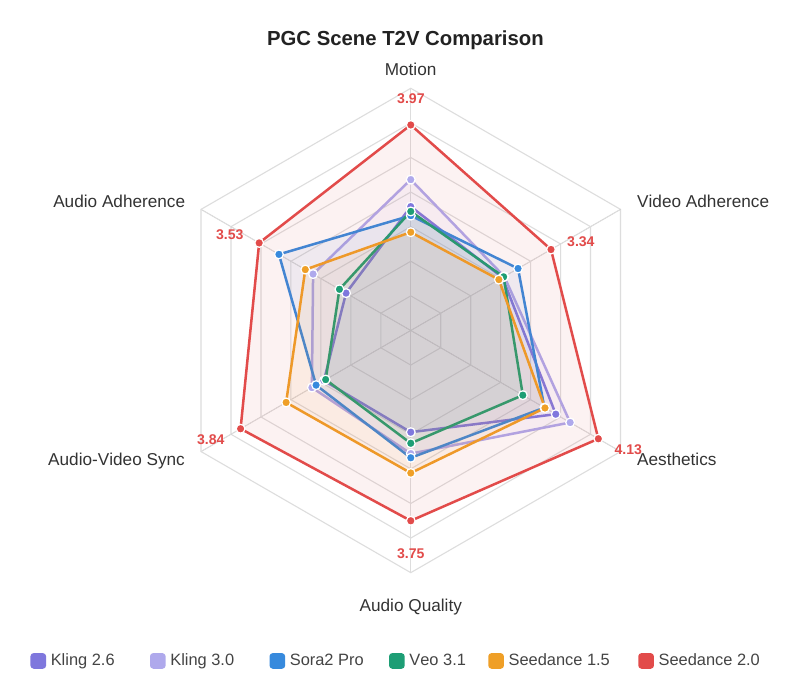}
        \caption{PGC Scene}
    \end{subfigure}
    \hfill
    \begin{subfigure}[b]{0.49\textwidth}
        \includegraphics[width=\textwidth]{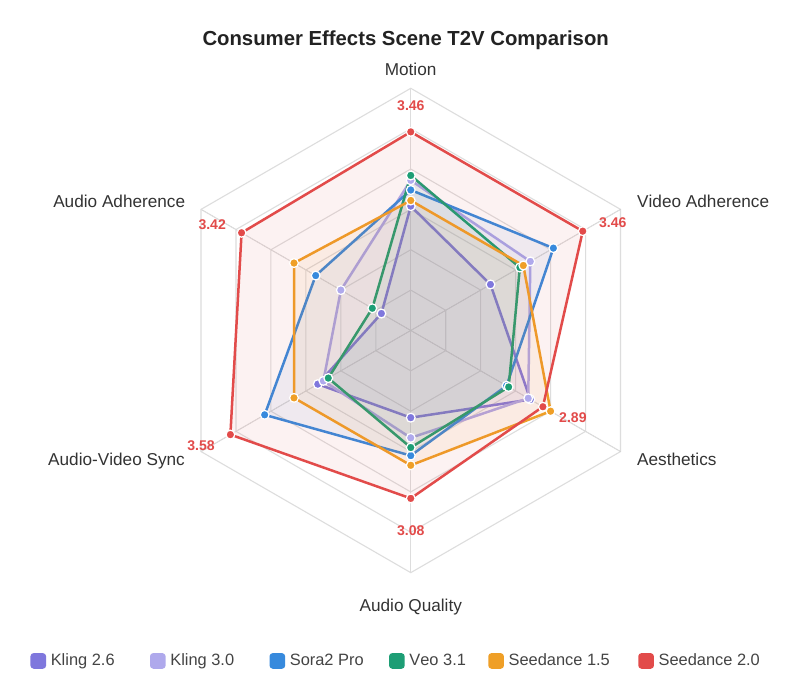}
        \caption{Consumer Effects Scene}
    \end{subfigure}

    \vspace{0.1em}

    \begin{subfigure}[b]{0.49\textwidth}
        \includegraphics[width=\textwidth]{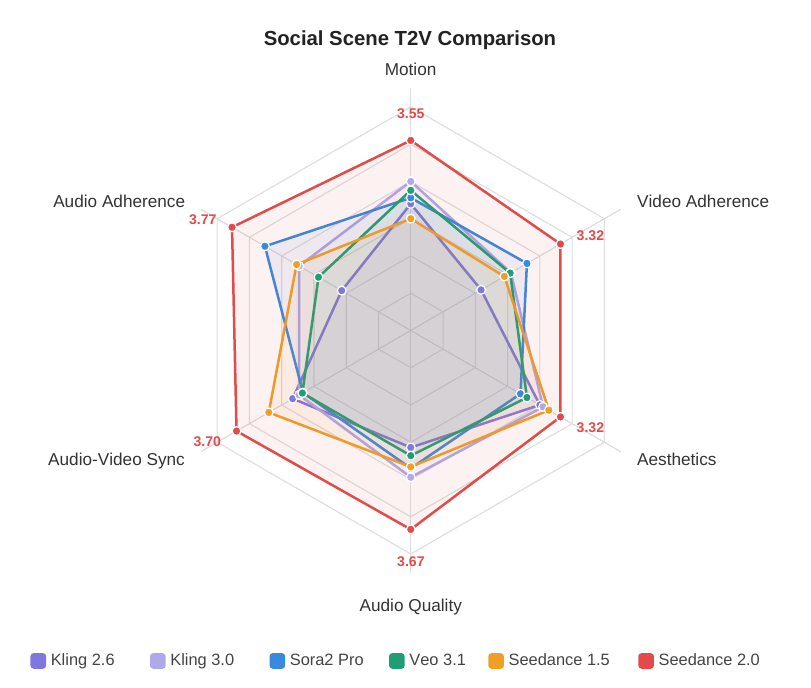}
        \caption{Social Scene}
    \end{subfigure}
    \hfill
    \begin{subfigure}[b]{0.49\textwidth}
        \includegraphics[width=\textwidth]{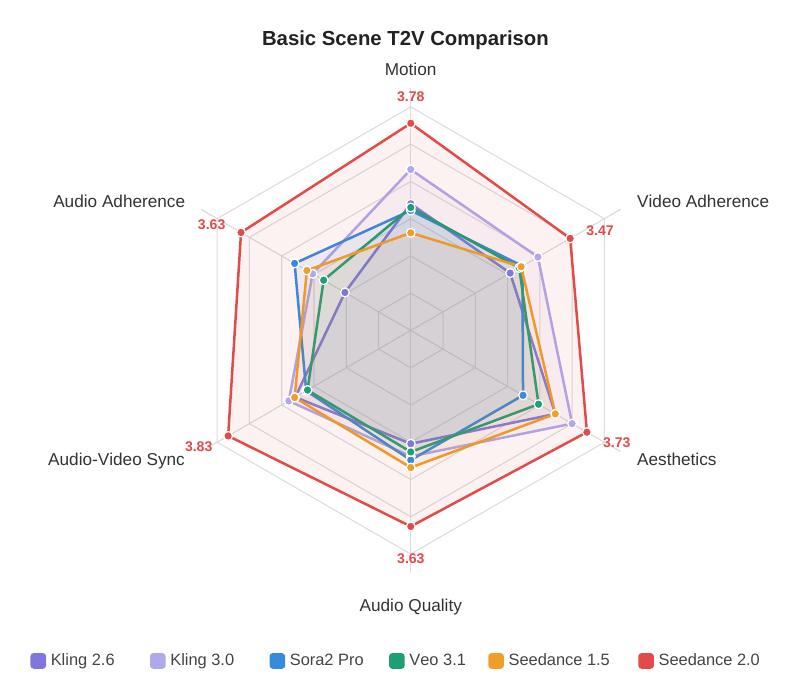}
        \caption{Basic Scene}
    \end{subfigure}

    \caption{T2V performance comparison across six scenarios.}
    \label{fig:t2v_comparison}
\end{figure}

\subsubsection{Motion Quality}
\input{tables/t2v/t2v_motion}
Seedance 2.0 leads on motion stability, editing rhythm, and multi-entity interaction, with far fewer cases of subject deformation and physically implausible motion than Seedance 1.5. Kling 3.0 ranks second but is limited on high-difficulty actions and complex camera work. Veo 3.1 and Sora 2 Pro handle basic motion reasonably well, yet fall short on fine-grained dynamics and long-take stability. The fine-grained breakdown in Table~\ref{tab:t2v_detailed} confirms this: Seedance 2.0 ranks first on 29 of 30 categories (tying with Kling 3.0 only on group coordinated motion), scoring 3.29--4.43. Multi-entity feature match (4.43), framing/composition (4.25), and editing rhythm (4.21) all exceed 4.0. Seedance 1.5 scored low on physical feedback (1.69), natural phenomena (2.00), and intense sports motion (2.00); Seedance 2.0 improves by over 1.5 points on each. Kling 3.0 does well on emotion \& expression (3.64), abstract challenges (3.57), and multi-entity feature match (3.43), but drops on intense sports motion (2.86), surreal motion (2.86), and special camera shots (3.08). Sora 2 Pro scores lowest on surreal motion (2.00) and intense sports motion (2.21); Veo 3.1 is weakest on multi-entity feature match (2.50) and group coordinated motion (2.33).

\subsubsection{Video Prompt Following}
\input{tables/t2v/t2v_video_adherence}
Compared to Seedance 1.5, version 2.0 improves on text rendering, physical phenomena, intent comprehension, and style following, with more precise action following and reasonable creative interpretation beyond the core instructions. Kling 3.0 carries over strengths from Kling 2.6 on emotional expression and physical feedback, but stays weak on text following. Veo 3.1's main shortcoming is poor text generation and instruction response. Sora 2 Pro handles most categories well with strong creative interpretation, leading on abstract challenges, but scores lowest on surreal motion, placing it in the second tier. Table~\ref{tab:t2v_video_adherence_detailed} shows Seedance 2.0 first on 27 of 30 categories, scoring 2.71--4.29. The largest gains over Seedance 1.5 appear on text-related categories---creative text (1.86 $\to$ 3.43), short text (2.00 $\to$ 3.57), text overlay (2.15 $\to$ 3.31)---and on physical phenomena (1.92 $\to$ 3.31) and natural phenomena (2.56 $\to$ 3.89). Counter-reality instructions (4.29) and emotion \& expression (4.00) are its two highest categories. Sora 2 Pro leads on abstract challenges (4.17 vs.\ 3.86) and framing/composition (3.50 vs.\ 3.13), but drops to 1.86 on surreal motion. Veo 3.1 leads on anthropomorphic motion (3.00) but scores below 2.2 on text overlay (2.17), short text (2.17), and creative text (1.67). Kling 3.0 scores 3.43 on emotion \& expression and 3.23 on physical feedback, but falls to 2.00 on text overlay and creative text.

\subsubsection{Video Aesthetics}
\input{tables/t2v/t2v_aesthetics}
Aesthetics is the most competitive dimension. Seedance 2.0 leads on visual effects, scene design, lighting and color, and realistic detail. Kling 3.0 is relatively strong on stylization and color expression, while other competitors are weaker on photorealism and fine detail. In Table~\ref{tab:t2v_aesthetics_detailed}, Seedance 2.0 ranks first or tied for first on 28 of 30 categories, scoring 2.79--4.14. Its highest scores are visual style (4.14), long script (4.14), framing/composition (4.13), and four categories at 4.00 (cinematic visual effects, editing rhythm, natural phenomena, multi-entity feature match). Seedance 2.0 does not lead on consumer visual effects (Seedance 1.5: 3.00 vs.\ 2.79) or surreal motion (Kling 3.0: 3.86 vs.\ 3.57), and ties on anthropomorphic motion (3.71, three-way) and advanced camera movement (3.54, tied with Kling 2.6). Kling 3.0 scores above 3.5 on 13 categories, with its best on surreal motion (3.86), same-type interaction (3.79), and framing/composition (3.75). Sora 2 Pro and Veo 3.1 lag: Sora 2 Pro drops below 2.5 on holidays (2.38), consumer visual effects (2.38), and natural phenomena (2.33); Veo 3.1 scores lowest on holidays (2.36), consumer visual effects (2.45), and multi-entity feature match (2.50).

\subsubsection{Audio Quality}
\input{tables/t2v/t2v_audio_quality}
Moving to audio, Seedance 2.0 improves over Seedance 1.5 on vocal and singing expressiveness, BGM-to-visual matching, and audio layering. Kling 3.0 regresses from Kling 2.6 on singing/rap and ambient background sound. Sora 2 Pro produces vivid audio with strength on singing, though limited to a narrow set of categories. Competitors broadly have muddy audio, noticeable noise, and weak layering, especially on complex sound effects and vocal clarity. Per Table~\ref{tab:t2v_audio_quality_detailed}, Seedance 2.0 ranks first on all 17 categories, scoring 2.82--4.17. English (4.17), voice + action interaction (4.00), minority languages (3.82), and ambient/background sound (3.78) are strongest. The improvement over Seedance 1.5 is largest on Chinese opera (2.50 $\to$ 3.75), English (3.00 $\to$ 4.17), and singing/rap (2.71 $\to$ 3.71). Kling 3.0 drops below Kling 2.6 on singing/rap (2.71 vs.\ 3.14) and ambient/background sound (2.33 vs.\ 2.78), despite improving elsewhere. Sora 2 Pro scores 3.67 on singing/rap (second only to Seedance 2.0) and 3.17 on voice + action interaction, but falls to 2.17 on Chinese opera and 2.44 on ambient sound. No competitor exceeds 3.2 on any category apart from Sora 2 Pro's singing/rap (3.67).

\subsubsection{Audio-Visual Sync}
\input{tables/t2v/t2v_audio_video_sync}
Lip synchronization and action-audio alignment are both strong for Seedance 2.0, with very few cases of delay or misalignment. Competitors commonly produce lip-speech mismatches and action-sound offsets, worse in fast dialogue and complex action scenes. In Table~\ref{tab:t2v_av_sync_detailed}, Seedance 2.0 ranks first on 16 of 17 categories (tying with Seedance 1.5 on off-screen voice at 2.86), scoring 2.86--4.17. English (4.17), singing/rap (4.14), dual-channel audio (4.00), and non-verbal voice (4.00) are strongest. The largest gains over Seedance 1.5 are Chinese multi-person dialogue (2.36 $\to$ 3.86), object interaction sound (2.65 $\to$ 3.82), and animal sound (2.79 $\to$ 3.93). No competitor reaches 3.5 on any category apart from Sora 2 Pro on singing/rap (3.50) and Seedance 1.5 on English (3.50). Veo 3.1 is weakest, dropping to 1.67 on spatial scene and 2.10 on Chinese multi-person dialogue. Kling 3.0 regresses from Kling 2.6 on singing/rap (2.57 vs.\ 3.00) and instruments \& audio (2.79 vs.\ 3.00), while improving on Chinese dialect (3.14 vs.\ 2.68).

\subsubsection{Audio Prompt Following}
\input{tables/t2v/t2v_audio_adherence}
Audio prompt following is the dimension where competitors score lowest. Seedance 2.0 is strong on complex audio instructions involving multilingual dialogue, dialect-specific speech, and diverse sound profiles such as animal vocalizations. Sora 2 Pro has an edge on instrument and natural sound effect following, while other models are weaker at generating specific timbres and language-accurate audio. Table~\ref{tab:t2v_audio_adherence_detailed} shows Seedance 2.0 first on 16 of 17 categories (tying with Kling 3.0 on off-screen voice at 3.14), scoring 2.91--4.25. English (4.25), instruments \& audio (3.89), ambient/background sound (3.89), voice + action interaction (3.86), and animal sound (3.86) are strongest. The largest gains over Seedance 1.5 are Chinese opera (1.75 $\to$ 3.50), singing/rap (2.14 $\to$ 3.71), and animal sound (2.50 $\to$ 3.86). Chinese dialect drops below 2.0 for five of six models, and Chinese opera below 2.4 for five of six. Sora 2 Pro is the strongest competitor, scoring 3.67 on singing/rap, 3.64 on English, 3.61 on instruments \& audio, and 3.50 on voice + action interaction, but falls to 1.86 on Chinese dialect. Kling 3.0 regresses from Kling 2.6 on object interaction sound (2.00 vs.\ 2.41) and dual-channel audio (2.57 vs.\ 2.71). Veo 3.1 scores below 2.0 on Chinese dialect (1.20), Chinese variety show voice (1.57), Chinese opera (1.29), and off-screen voice (1.83).

\subsection{Image-to-Video Evaluation on SeedVideoBench 2.0}
\subsubsection{Overall Results}
\input{tables/i2v/i2v_overall_score}
Table~\ref{tab:i2v_mos} summarizes the I2V results. Seedance 2.0 ranks first on all six dimensions, scoring 3.31--3.70, while no competitor exceeds 3.18. The three video dimensions show 3.35 (motion quality), 3.46 (video prompt following), and 3.31 (image preservation). Source image preservation is the tightest race---Kling 3.0 trails by only 0.13---while motion quality shows a 0.55-point gap to the runner-up. The three audio dimensions are where competitors fall furthest behind: Seedance 2.0 scores 3.61, 3.54, and 3.70, while Kling 2.6 (2.21) and Wan 2.6~\cite{wan2p6} (2.18--2.55) sit well below 3.0. Seedance 1.5 Pro is second on audio (3.07, 2.95, 3.10) but still trails by 0.54--0.60. Audio prompt following (3.70) is Seedance 2.0's highest I2V score. A two-tier pattern is clear: Seedance 2.0 leads on both video and audio, while competitors are weaker on audio than on video.

\input{tables/i2v/i2v_overall_analysis}
Table~\ref{tab:i2v_rates} breaks this down further. Seedance 2.0 is the only model with usability above 87\% on all six dimensions. On motion quality, its 43.88\% satisfaction rate is over 3$\times$ the runner-up Kling 3.0 (12.00\%); on video prompt following, 47.48\% vs.\ Veo 3.1's 20.54\%. Source image preservation is again closest: 91.37\% usability vs.\ Kling 3.0's 90.55\%, though the satisfaction gap is wider (38.85\% vs.\ 27.27\%). The audio contrast is sharper. On audio quality, Seedance 2.0 reaches 97.42\% usability and 57.08\% satisfaction; Kling 2.6 and Wan 2.6 have usability below 28\%, meaning most of their audio is rated unacceptable. On audio prompt following, Seedance 2.0's 63.52\% satisfaction is 1.7$\times$ Seedance 1.5 Pro's 37.77\% and over 10$\times$ Kling 2.6's 5.70\%.

Beyond the scores, human evaluation of generated videos surfaces additional patterns. Seedance 2.0 produces dynamic motion with a clear sense of momentum---combat and dance sequences mix slow-motion highlights with fast action in ways competitors do not, and facial expressions and gaze are more vivid than Seedance 1.5 Pro. Camera work tracks subject motion closely with varied angles and smooth push/pull transitions. First- and third-person game-following perspectives with handheld breathing effects are new to this version and add immersion. The model handles special art styles (felt, oil painting, Chinese gongbi) without breaking visual coherence, matching subject motion to the referenced style. Realistic and 3D visual effects render fluidly. On the audio side, dialogue voices carry emotional nuance in both Chinese and non-Chinese languages. Voice, sound effects, and audio are well-layered---outputs sound like composed audio rather than isolated tracks stacked together. Common Chinese dialects (Sichuan, Northeastern, Cantonese) come through accurately. Singing, rap, and instrumental audio across languages are strong, with melodies that fit the prompt context.

\subsubsection{Detailed Visual Evaluation Results}
\input{tables/i2v/i2v_analysis_visual}

\subsubsection{Detailed Audio Evaluation Results}
\input{tables/i2v/i2v_analysis_audio}

\subsection{Reference-to-Video Evaluation on SeedVideoBench 2.0}
\subsubsection{Quantitative Results}
\input{tables/r2v/r2v_score}
Seedance 2.0 leads all evaluated models on R2V, outperforming Kling 3 Omni~\cite{kling3.0}, Kling O1~\cite{klingO1}, and Vidu Q2 Pro~\cite{viduq2} across every dimension. It supports more multi-modal task types with higher accuracy---beyond subject, style, motion reference, and video editing supported by competitors, Seedance 2.0 also handles creative and visual-effects reference, video continuation and extension, and combined tasks such as motion reference + subject reference, with fewer issues of missing or confused input materials. Reference alignment is best across subject appearance and voice, motion, and style, with clear advantages in subject identity and style preservation. Motion quality---vividness, physical plausibility, and structural stability---is uniformly stronger, with stability as the largest lead. Prompt following is also strongest, particularly on long-text and complex narratives, visual effects, dialogue, text rendering, and open-ended instructions. Table~\ref{tab:r2v_eval} quantifies this: Seedance 2.0 ranks first on all five dimensions, scoring 2.50 and 2.52 on multimodal task following and prompt following (1--3 scale), and 3.54, 3.03, 3.24 on editing consistency, reference alignment, and motion quality (1--5 scale). The gaps are smallest on editing consistency (Kling 3.0 trails by 0.17) and largest on motion quality (0.86--0.94 behind across all competitors) and reference alignment (0.66--1.24 behind). Vidu Q2 Pro scores lowest on three of five dimensions. Kling 3.0 and Kling O1 trade second and third depending on the dimension, but neither approaches Seedance 2.0 on motion quality or prompt following.

\input{tables/r2v/r2v_support_matrix}
Table~\ref{tab:r2v_task_support} compares multi-modal task support. Seedance 2.0 supports 20 of 22 input modalities---the broadest of any model. The two unsupported tasks (subject audio-visual + audio reference, video audio editing) are unsupported by every model. Three task groups are exclusive to Seedance 2.0: all three visual effects / creative reference variants and all four continuation / extension variants, totaling 7 tasks no competitor can handle. Kling 3 Omni supports 9 of 22, lacking style reference, visual effects / creative reference, and continuation / extension. Vidu Q2 Pro supports 13 of 22, covering style reference but missing visual effects / creative reference and continuation / extension. Kling O1 is the most limited at 10 of 22, additionally lacking video-based subject reference and audio-visual inputs.

\input{tables/r2v/r2v_analysis}

\subsection{Visualization Results}
\begin{figure}[H]
    \centering
    \includegraphics[width=1.0\textwidth]{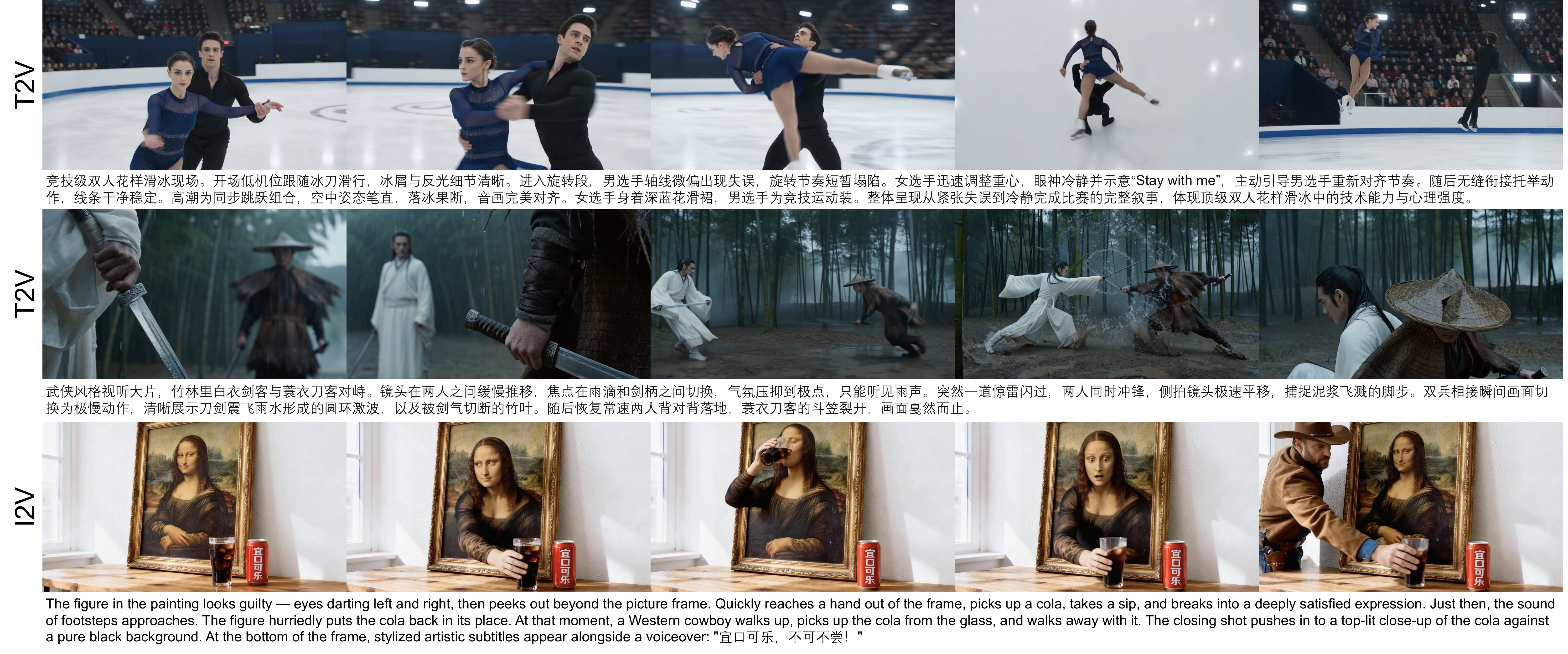}
    \vspace{-0.8cm}
    \caption{Visualization of text-to-video (T2V) and image-to-video (I2V) generation.}
    \label{fig:t2v_i2v_visualization}
\end{figure}

Leveraging significant advancements in foundational capabilities and Multimodal Task Performance, Seedance 2.0 will deliver an entirely new creative experience for users. It is capable of synthesizing temporally precise and complex interactive scenes with high fidelity. As shown in the first and second row of Figure~\ref{fig:t2v_i2v_visualization}, the generation process maintains exceptional motion quality by strictly adhering to real-world physical laws of motion, avoiding physical anomalies commonly observed in earlier AI-generated videos. For example, in the skating scenario, the model effectively renders a series of highly demanding maneuvers—such as synchronized take-offs, mid-air rotations, and precise landings. In response to the diversified demands of video content production, Seedance 2.0 demonstrates exceptional scenario adaptability in the third row of Figure~\ref{fig:t2v_i2v_visualization}. Whether applied to commercial advertising, television visual effects, video game animation, or explainer videos, the model consistently delivers high-quality generation results and robust multimodal task performance. By replacing complex visual effects pipelines and live-action workflows with AI-driven generation, Seedance 2.0 significantly reduces the production costs of professional audio and video content, shortens production cycles, and empowers both creators and enterprises to more effectively realize their creative visions.

%% file: tables/t2v/t2v_overall.tex
\begin{table}[H]
  \centering
  \caption{T2V overall evaluation results across six dimensions (Rating from 1--5).}
  \vspace{-0.3cm}
  \label{tab:t2v_mos}
  \begin{tabular}{lcccccc}
  \toprule
  \textbf{Model} & \textbf{Motion} & \textbf{\makecell{Video Prompt\\Following}} & \textbf{Aesthetics} & \textbf{\makecell{Audio\\Quality}} & \textbf{\makecell{Audio-Visual\\Sync}} & \textbf{\makecell{Audio Prompt\\Following}} \\
  \midrule
  Kling 2.6~\cite{kling2p6}      & 2.72 & 2.39 & 3.21 & 2.46 & 2.67 & 2.00 \\
  Kling 3.0~\cite{kling3.0}      & \underline{3.10} & 2.78 & \underline{3.36} & 2.74 & 2.78 & 2.54 \\
  Sora2 Pro~\cite{sora2}      & 2.69 & \underline{2.81} & 2.82 & 2.76 & 2.65 & \underline{2.92} \\
  Veo3.1~\cite{veo3p1}         & 2.73 & 2.59 & 2.88 & 2.62 & 2.54 & 2.24 \\
  Seedance 1.5~\cite{seedance1p5}   & 2.39 & 2.59 & 3.19 & \underline{2.88} & \underline{2.91} & 2.69 \\
  Seedance 2.0   & \textbf{3.75} & \textbf{3.43} & \textbf{3.67} & \textbf{3.63} & \textbf{3.75} & \textbf{3.56} \\
  \bottomrule
  \end{tabular}%
\end{table}

%% file: tables/t2v/t2v_usability_rate.tex
\begin{table}[H]
  \centering
  \caption{T2V usability, satisfaction, and delight rates across six evaluation dimensions.}
  \vspace{-0.3cm}
  \label{tab:t2v_rates}
  \fittowidth{%
  \begin{tabular}{l cccccc}
  \toprule
  & \textbf{\makecell{Motion\\Quality}} & \textbf{\makecell{Video Prompt\\Following}} & \textbf{Aesthetics} & \textbf{\makecell{Audio\\Quality}} & \textbf{\makecell{Audio-Visual\\Sync}} &
\textbf{\makecell{Audio Prompt\\Following}} \\
  \midrule
  \multicolumn{7}{c}{\textbf{\textit{Usability Rate (score $\geq$ 3)}}} \\
  \midrule
  Kling 2.6~\cite{kling2p6}      & 70.55\% & 41.72\% & 90.80\% & 45.98\% & 58.04\% & 29.02\% \\
  Kling 3.0~\cite{kling3.0}      & \underline{82.82\%} & 58.90\% & 91.10\% & 66.52\% & 66.96\% & 54.02\% \\
  Sora2 Pro~\cite{sora2}      & 65.08\% & \underline{62.54\%} & 63.17\% & 66.82\% & 59.35\% & \underline{63.08\%} \\
  Veo3.1~\cite{veo3p1}         & 67.13\% & 53.63\% & 74.39\% & 60.51\% & 54.36\% & 37.44\% \\
  Seedance 1.5~\cite{seedance1p5}   & 46.93\% & 54.29\% & \textbf{96.93\%} & \underline{82.59\%} & \underline{69.64\%} & 56.70\% \\
  Seedance 2.0   & \textbf{97.55\%} & \textbf{84.97\%} & \underline{96.32\%} & \textbf{93.75\%} & \textbf{93.30\%} & \textbf{83.93\%} \\
  \midrule
  \multicolumn{7}{c}{\textbf{\textit{Satisfaction Rate (score $\geq$ 4)}}} \\
  \midrule
  Kling 2.6~\cite{kling2p6}      & 3.99\%  & 9.20\%  & 29.75\% & 2.75\%  & 18.75\% & 8.04\% \\
  Kling 3.0~\cite{kling3.0}      & \underline{28.22\%} & 21.47\% & \underline{43.56\%} & 9.59\%  & 15.18\% & 17.86\% \\
  Sora2 Pro~\cite{sora2}      & 6.98\%  & \underline{22.54\%} & 20.00\% & \underline{9.81\%}  & 19.16\% & \underline{31.78\%} \\
  Veo3.1~\cite{veo3p1}         & 6.57\%  & 12.11\% & 13.84\% & 4.10\%  & 13.33\% & 14.87\% \\
  Seedance 1.5~\cite{seedance1p5}   & 1.23\%  & 12.27\% & 21.78\% & 5.36\%  & \underline{25.45\%} & 25.45\% \\
  Seedance 2.0   & \textbf{67.18\%} & \textbf{51.23\%} & \textbf{61.66\%} & \textbf{62.05\%} & \textbf{68.30\%} & \textbf{57.94\%} \\
  \midrule
  \multicolumn{7}{c}{\textbf{\textit{Delight Rate (score $=$ 5)}}} \\
  \midrule
  Kling 2.6~\cite{kling2p6}      & 0.00\%  & 0.31\%  & \underline{1.53\%}  & 0.00\%  & 1.34\%  & 0.45\% \\
  Kling 3.0~\cite{kling3.0}      & \underline{0.61\%}  & 2.76\%  & \underline{1.53\%}  & 0.00\%  & 1.34\%  & 2.23\% \\
  Sora2 Pro~\cite{sora2}      & 0.00\%  & \underline{2.86\%}  & 0.63\%  & 0.00\%  & 0.47\%  & \underline{11.68\%} \\
  Veo3.1~\cite{veo3p1}         & 0.00\%  & 0.69\%  & 0.35\%  & 0.00\%  & 0.00\%  & 0.51\% \\
  Seedance 1.5~\cite{seedance1p5}   & 0.00\%  & 0.31\%  & 0.00\%  & 0.00\%  & \underline{1.79\%}  & 1.79\% \\
  Seedance 2.0   & \textbf{10.43\%} & \textbf{8.28\%} & \textbf{9.20\%} & \textbf{6.70\%} & \textbf{13.84\%} & \textbf{26.92\%} \\
  \bottomrule
  \end{tabular}%
  }
  \vspace{-0.3cm}
\end{table}

%% file: tables/t2v/t2v_motion.tex
\begin{table}[H]
  \centering
  \caption{T2V detailed motion evaluation results across fine-grained categories. Rating from 1 to 5, with higher scores indicating better performance.}
  \label{tab:t2v_detailed}
  \fittowidth{%
  \begin{tabular}{lcccccc}
  \toprule
  \textbf{Category} & \textbf{Kling 2.6}~\cite{kling2p6} & \textbf{Kling 3.0}~\cite{kling3.0} & \textbf{Sora2 Pro}~\cite{sora2} & \textbf{Veo3.1}~\cite{veo3p1} & \textbf{Seedance 1.5}~\cite{seedance1p5} & \textbf{Seedance 2.0} \\
  \midrule
  Holidays / Festivals          & 2.57 & 2.57 & 2.69 & \underline{3.00} & 2.71 & \textbf{3.29} \\
  Consumer Visual Effects       & 2.64 & \underline{3.00} & 2.77 & 2.82 & 2.43 & \textbf{3.71} \\
  Counter-Reality Instructions  & 2.86 & \underline{3.00} & 2.67 & \underline{3.00} & \underline{3.00} & \textbf{3.71} \\
  Cinematic Visual Effects           & \underline{3.00} & \underline{3.00} & 2.57 & 2.82 & 2.57 & \textbf{3.79} \\
  Same-Type Interaction         & 2.57 & \underline{3.29} & 2.64 & 2.62 & 2.29 & \textbf{3.79} \\
  Cross-Type Interaction        & 2.79 & \underline{3.14} & 2.79 & 2.77 & 2.43 & \textbf{3.57} \\
  Group Coordinated Motion      & \underline{2.71} & \textbf{3.29} & 2.57 & 2.33 & 2.57 & \textbf{3.29} \\
  Advanced Camera Movement      & 3.00 & \underline{3.23} & 2.58 & 2.80 & 2.31 & \textbf{3.77} \\
  Special Camera Shots          & 2.85 & \underline{3.08} & 2.75 & 2.64 & 2.08 & \textbf{3.92} \\
  Editing Rhythm                & 2.86 & \underline{3.14} & 2.93 & 2.69 & 2.43 & \textbf{4.21} \\
  Combined Shot Instructions    & 2.86 & \underline{3.00} & 2.29 & 2.67 & 2.29 & \textbf{3.86} \\
  Physical Feedback             & 2.69 & \underline{3.00} & 2.69 & 2.83 & 1.69 & \textbf{3.46} \\
  Physical Phenomena            & 2.46 & \underline{3.08} & 2.77 & 2.58 & 2.23 & \textbf{3.38} \\
  Natural Phenomena             & 2.56 & \underline{3.11} & 2.78 & 2.57 & 2.00 & \textbf{3.78} \\
  Text Overlay                  & 2.46 & \underline{2.85} & 2.64 & 2.67 & 2.38 & \textbf{3.69} \\
  Short Text                    & 3.00 & \underline{3.14} & 3.00 & 2.67 & 2.57 & \textbf{3.71} \\
  Creative Text                 & 2.43 & \underline{2.86} & \underline{2.86} & 2.83 & 2.57 & \textbf{3.57} \\
  Long Script                   & 3.00 & \underline{3.29} & 2.57 & 2.83 & 2.57 & \textbf{3.57} \\
  Abstract Challenges           & 3.00 & \underline{3.57} & 3.33 & 2.57 & 2.57 & \textbf{4.00} \\
  Multi-Entity Feature Match    & 3.00 & \underline{3.43} & 3.00 & 2.50 & 2.29 & \textbf{4.43} \\
  Knowledge Assessment          & 2.62 & \underline{3.38} & 2.62 & 2.91 & 2.62 & \textbf{3.69} \\
  Compound Multi-Instructions   & 2.57 & \underline{2.86} & 2.29 & 2.40 & 2.57 & \textbf{3.71} \\
  Surreal Motion                & 2.43 & \underline{2.86} & 2.00 & 2.43 & 2.43 & \textbf{3.71} \\
  Intense Sports Motion         & 2.43 & \underline{2.86} & 2.21 & 2.43 & 2.00 & \textbf{3.79} \\
  Fine Hand Motion              & 2.64 & \underline{3.00} & 2.57 & 2.69 & 2.36 & \textbf{3.71} \\
  Anthropomorphic Motion        & 2.57 & \underline{2.71} & 2.43 & \underline{2.71} & 2.29 & \textbf{3.29} \\
  Emotion \& Expression         & 2.86 & \underline{3.64} & 2.93 & 3.00 & 2.64 & \textbf{4.00} \\
  Visual Style                  & 2.71 & \underline{3.14} & 2.77 & 2.62 & 2.50 & \textbf{4.00} \\
  Lighting \& Color Tone        & 2.71 & \underline{3.29} & 2.92 & 2.93 & 2.36 & \textbf{3.71} \\
  Framing / Composition         & 3.13 & \underline{3.38} & 3.00 & 3.25 & 2.63 & \textbf{4.25} \\
  \bottomrule
  \end{tabular}%
  }
\end{table}

%% file: tables/t2v/t2v_video_adherence.tex
\begin{table}[ht!]
  \centering
  \caption{T2V detailed Video Prompt Following evaluation results across fine-grained categories. Rating
  from 1 to 5, with higher scores indicating better performance.}
  \label{tab:t2v_video_adherence_detailed}
  \fittowidth{%
  \begin{tabular}{lcccccc}
  \toprule
  \textbf{Category} & \textbf{Kling 2.6}~\cite{kling2p6} & \textbf{Kling 3.0}~\cite{kling3.0} & \textbf{Sora2 Pro}~\cite{sora2} & \textbf{Veo3.1}~\cite{veo3p1} & \textbf{Seedance 1.5}~\cite{seedance1p5} & \textbf{Seedance 2.0} \\
  \midrule
  Holidays / Festivals          & 2.14 & 2.50 & \underline{3.08} & 2.57 & 2.50 & \textbf{3.57} \\
  Consumer Visual Effects       & 2.14 & 2.93 & \underline{3.00} & 2.55 & 2.79 & \textbf{3.36} \\
  Counter-Reality Instructions  & 2.43 & 3.00 & 3.17 & \underline{3.40} & 3.00 & \textbf{4.29} \\
  Cinematic Visual Effects           & 2.71 & 3.00 & 2.50 & \underline{3.09} & 3.00 & \textbf{3.64} \\
  Same-Type Interaction         & 2.29 & \underline{3.07} & 2.36 & 2.62 & 2.36 & \textbf{3.64} \\
  Cross-Type Interaction        & 2.64 & \underline{2.79} & 2.57 & 2.69 & \underline{2.79} & \textbf{3.50} \\
  Group Coordinated Motion      & 2.86 & \underline{3.14} & 2.86 & 2.50 & 2.57 & \textbf{3.86} \\
  Advanced Camera Movement      & 2.69 & \underline{2.85} & 2.50 & 2.40 & 2.77 & \textbf{3.46} \\
  Special Camera Shots          & 2.00 & 2.62 & \underline{2.75} & 2.27 & 2.46 & \textbf{3.00} \\
  Editing Rhythm                & 2.57 & 2.50 & \underline{2.93} & 2.62 & 2.43 & \textbf{3.14} \\
  Combined Shot Instructions    & \underline{2.71} & 2.57 & 2.29 & 2.33 & 2.43 & \textbf{3.29} \\
  Physical Feedback             & 3.08 & \underline{3.23} & 2.77 & 2.75 & 2.46 & \textbf{3.62} \\
  Physical Phenomena            & 2.23 & \underline{2.77} & \underline{2.77} & 2.33 & 1.92 & \textbf{3.31} \\
  Natural Phenomena             & 2.33 & 3.11 & \underline{3.22} & 2.71 & 2.56 & \textbf{3.89} \\
  Text Overlay                  & 1.85 & 2.00 & \underline{2.91} & 2.17 & 2.15 & \textbf{3.31} \\
  Short Text                    & 1.86 & 2.29 & \underline{3.33} & 2.17 & 2.00 & \textbf{3.57} \\
  Creative Text                 & 1.71 & 2.00 & \underline{3.00} & 1.67 & 1.86 & \textbf{3.43} \\
  Long Script                   & 2.00 & 2.86 & \underline{3.00} & 2.67 & 2.43 & \textbf{3.29} \\
  Abstract Challenges           & 2.00 & 3.14 & \textbf{4.17} & 2.86 & 2.86 & \underline{3.86} \\
  Multi-Entity Feature Match    & 2.14 & \underline{3.14} & 3.00 & 2.17 & 2.43 & \textbf{3.86} \\
  Knowledge Assessment          & 2.23 & 2.54 & 2.77 & \underline{3.00} & \underline{3.00} & \textbf{3.23} \\
  Compound Multi-Instructions   & 2.71 & \underline{3.14} & 2.14 & 2.40 & 2.57 & \textbf{3.71} \\
  Surreal Motion                & \underline{2.43} & \textbf{2.71} & 1.86 & 2.00 & 2.14 & \textbf{2.71} \\
  Intense Sports Motion         & 2.64 & \underline{2.79} & 2.36 & 2.43 & 2.71 & \textbf{3.21} \\
  Fine Hand Motion              & 2.29 & \underline{2.93} & 2.86 & 2.38 & 2.57 & \textbf{3.50} \\
  Anthropomorphic Motion        & 2.00 & 2.14 & \underline{2.86} & \textbf{3.00} & 2.57 & \underline{2.86} \\
  Emotion \& Expression         & 2.64 & \underline{3.43} & 3.21 & 3.40 & 2.93 & \textbf{4.00} \\
  Visual Style                  & 2.36 & 2.21 & \underline{2.62} & 2.54 & 2.14 & \textbf{2.93} \\
  Lighting \& Color Tone        & 2.50 & 2.79 & \underline{3.00} & 2.79 & 2.50 & \textbf{3.21} \\
  Framing / Composition         & 2.88 & \underline{3.25} & \textbf{3.50} & 3.00 & 2.63 & 3.13 \\
  \bottomrule
  \end{tabular}%
  }
\end{table}

%% file: tables/t2v/t2v_aesthetics.tex
\begin{table}[ht!]
  \centering
  \caption{T2V detailed aesthetics evaluation results across fine-grained categories. Rating from 1
  to 5, with higher scores indicating better performance.}
  \label{tab:t2v_aesthetics_detailed}
  \fittowidth{%
  \begin{tabular}{lcccccc}
  \toprule
  \textbf{Category} & \textbf{Kling 2.6}~\cite{kling2p6} & \textbf{Kling 3.0}~\cite{kling3.0} & \textbf{Sora2 Pro}~\cite{sora2} & \textbf{Veo3.1}~\cite{veo3p1} & \textbf{Seedance 1.5}~\cite{seedance1p5} & \textbf{Seedance 2.0} \\
  \midrule
  Holidays / Festivals          & \underline{2.71} & \underline{2.71} & 2.38 & 2.36 & \textbf{3.00} & \textbf{3.00} \\
  Consumer Visual Effects       & 2.71 & 2.64 & 2.38 & 2.45 & \textbf{3.00} & \underline{2.79} \\
  Counter-Reality Instructions  & 3.43 & \underline{3.57} & 3.00 & 3.20 & 3.29 & \textbf{3.86} \\
  Cinematic Visual Effects           & \underline{3.64} & \underline{3.64} & 3.00 & 2.73 & 3.14 & \textbf{4.00} \\
  Same-Type Interaction         & 3.29 & \underline{3.79} & 2.79 & 2.85 & 3.29 & \textbf{3.86} \\
  Cross-Type Interaction        & \underline{3.36} & 3.14 & 3.14 & 3.15 & \underline{3.36} & \textbf{3.43} \\
  Group Coordinated Motion      & 2.86 & \underline{3.00} & 2.43 & 2.50 & \underline{3.00} & \textbf{3.29} \\
  Advanced Camera Movement      & \textbf{3.54} & \underline{3.38} & 2.83 & 3.00 & 2.92 & \textbf{3.54} \\
  Special Camera Shots          & 3.08 & \underline{3.46} & 2.67 & 3.09 & 3.15 & \textbf{3.85} \\
  Editing Rhythm                & 3.21 & \underline{3.57} & 3.29 & 3.08 & 3.29 & \textbf{4.00} \\
  Combined Shot Instructions    & \underline{3.29} & \textbf{3.57} & 3.00 & 2.67 & \underline{3.29} & \textbf{3.57} \\
  Physical Feedback             & 3.31 & 3.31 & 2.46 & \underline{3.33} & 3.23 & \textbf{3.54} \\
  Physical Phenomena            & 3.00 & \underline{3.23} & 2.77 & 2.92 & 3.00 & \textbf{3.54} \\
  Natural Phenomena             & 3.33 & \underline{3.67} & 2.33 & 2.57 & 3.00 & \textbf{4.00} \\
  Text Overlay                  & 2.92 & \underline{3.15} & 3.00 & 2.75 & \underline{3.15} & \textbf{3.31} \\
  Short Text                    & \underline{3.43} & 3.14 & 3.00 & 3.00 & 3.00 & \textbf{3.86} \\
  Creative Text                 & 2.57 & 3.00 & 2.86 & \underline{3.17} & 3.00 & \textbf{3.57} \\
  Long Script                   & 3.29 & \underline{3.57} & 2.86 & 3.17 & 3.14 & \textbf{4.14} \\
  Abstract Challenges           & 3.29 & \underline{3.57} & 3.17 & 2.57 & 3.29 & \textbf{3.86} \\
  Multi-Entity Feature Match    & \underline{3.71} & 3.43 & 3.14 & 2.50 & 3.43 & \textbf{4.00} \\
  Knowledge Assessment          & 3.23 & \underline{3.31} & 2.62 & 3.09 & \underline{3.31} & \textbf{3.54} \\
  Compound Multi-Instructions   & 3.29 & \underline{3.57} & 2.86 & 2.80 & 3.29 & \textbf{3.86} \\
  Surreal Motion                & 3.14 & \textbf{3.86} & 2.86 & 2.57 & \underline{3.57} & \underline{3.57} \\
  Intense Sports Motion         & 3.07 & \underline{3.21} & 2.64 & 2.71 & 3.00 & \textbf{3.79} \\
  Fine Hand Motion              & \textbf{3.43} & 2.93 & 2.57 & 3.00 & \underline{3.21} & \textbf{3.43} \\
  Anthropomorphic Motion        & \textbf{3.71} & \textbf{3.71} & 3.29 & 3.29 & \underline{3.43} & \textbf{3.71} \\
  Emotion \& Expression         & 3.21 & \underline{3.71} & 2.93 & 3.10 & 3.29 & \textbf{3.86} \\
  Visual Style                  & 3.29 & \underline{3.50} & 3.15 & 2.92 & 3.21 & \textbf{4.14} \\
  Lighting \& Color Tone        & 3.00 & \underline{3.50} & 2.62 & 2.79 & 3.36 & \textbf{3.86} \\
  Framing / Composition         & 3.38 & \underline{3.75} & 3.38 & 2.88 & 3.25 & \textbf{4.13} \\
  \bottomrule
  \end{tabular}%
  }
\end{table}

%% file: tables/t2v/t2v_audio_quality.tex
\begin{table}[ht!]
  \centering
  \caption{T2V detailed audio quality evaluation results across fine-grained categories. Rating from
   1 to 5, with higher scores indicating better performance.}
  \label{tab:t2v_audio_quality_detailed}
  \fittowidth{%
  \begin{tabular}{lcccccc}
  \toprule
  \textbf{Category} & \textbf{Kling 2.6}~\cite{kling2p6} & \textbf{Kling 3.0}~\cite{kling3.0} & \textbf{Sora2 Pro}~\cite{sora2} & \textbf{Veo3.1}~\cite{veo3p1} & \textbf{Seedance 1.5}~\cite{seedance1p5} & \textbf{Seedance 2.0} \\
  \midrule
  Chinese Dialect / Accent        & 2.05 & \underline{2.41} & 2.29 & 2.10 & 2.32 & \textbf{2.82} \\
  Chinese Multi-Person Dialogue   & 2.36 & 2.93 & 2.79 & 2.20 & \underline{3.00} & \textbf{3.71} \\
  Chinese Variety Show Voice      & 2.14 & 2.57 & \underline{2.71} & 2.14 & 2.57 & \textbf{3.14} \\
  Chinese Opera                   & 2.13 & \underline{2.88} & 2.17 & 2.00 & 2.50 & \textbf{3.75} \\
  English                         & 3.08 & \underline{3.17} & 2.82 & 3.10 & 3.00 & \textbf{4.17} \\
  Minority Languages              & 2.03 & 2.59 & 2.85 & 2.89 & \underline{3.09} & \textbf{3.82} \\
  Singing / Rap                   & 3.14 & 2.71 & \underline{3.67} & 3.00 & 2.71 & \textbf{3.71} \\
  Spatial Scene                   & 2.57 & \underline{3.14} & 2.71 & 2.67 & 2.86 & \textbf{3.43} \\
  Off-Screen Voice                & 2.29 & \underline{3.00} & \underline{3.00} & 2.50 & \underline{3.00} & \textbf{3.29} \\
  Non-Verbal Voice                & 2.44 & 2.44 & \underline{2.78} & 2.56 & 2.67 & \textbf{3.56} \\
  Voice + Action Interaction      & 2.71 & 3.14 & \underline{3.17} & 2.67 & 3.00 & \textbf{4.00} \\
  Object Interaction Sound        & 2.59 & 2.47 & 2.65 & 2.76 & \underline{3.06} & \textbf{3.76} \\
  Animal Sound                    & 2.36 & 2.57 & 2.54 & 2.57 & \underline{2.79} & \textbf{3.57} \\
  Ambient / Background Sound      & 2.78 & 2.33 & 2.44 & 2.63 & \underline{3.00} & \textbf{3.78} \\
  Special Effects (ASMR, etc.)    & 2.59 & 2.76 & \underline{3.12} & 2.79 & 3.00 & \textbf{3.71} \\
  Instruments \& Audio            & 2.79 & \underline{3.00} & 2.78 & 2.89 & 2.95 & \textbf{3.68} \\
  Dual-Channel Audio              & 3.00 & 3.00 & 2.57 & 2.50 & \underline{3.14} & \textbf{3.43} \\
  \bottomrule
  \end{tabular}%
  }
\end{table}

%% file: tables/t2v/t2v_audio_video_sync.tex
\begin{table}[ht!]
         \centering
         \caption{T2V detailed audio-visual synchronization evaluation results across fine-grained
         categories. Rating from 1 to 5, with higher scores indicating better performance.}
         \label{tab:t2v_av_sync_detailed}
         \fittowidth{%
         \begin{tabular}{lcccccc}
         \toprule
         \textbf{Category} & \textbf{Kling 2.6}~\cite{kling2p6} & \textbf{Kling 3.0}~\cite{kling3.0} & \textbf{Sora2 Pro}~\cite{sora2} & \textbf{Veo3.1}~\cite{veo3p1} & \textbf{Seedance 1.5}~\cite{seedance1p5} & \textbf{Seedance 2.0} \\
         \midrule
         Chinese Dialect / Accent        & 2.68 & \underline{3.14} & 2.67 & 2.50 & 3.00 & \textbf{3.64} \\
         Chinese Multi-Person Dialogue   & 2.36 & \underline{2.93} & 2.64 & 2.10 & 2.36 & \textbf{3.86} \\
         Chinese Variety Show Voice      & 2.29 & 2.71 & 2.29 & 2.43 & \underline{2.86} & \textbf{3.14} \\
         Chinese Opera                   & 2.38 & 2.63 & 2.50 & \underline{2.71} & 2.63 & \textbf{3.63} \\
         English                         & 2.83 & 3.00 & 3.00 & 2.40 & \underline{3.50} & \textbf{4.17} \\
         Minority Languages              & 2.88 & \underline{2.97} & 2.53 & 2.68 & 2.74 & \textbf{3.88} \\
         Singing / Rap                   & 3.00 & 2.57 & \underline{3.50} & 3.00 & 3.29 & \textbf{4.14} \\
         Spatial Scene                   & 2.71 & \underline{3.14} & 2.43 & 1.67 & 3.00 & \textbf{3.86} \\
         Off-Screen Voice                & 2.29 & \underline{2.43} & 2.33 & 2.33 & \textbf{2.86} & \textbf{2.86} \\
         Non-Verbal Voice                & 2.67 & 2.56 & 2.44 & 2.56 & \underline{2.89} & \textbf{4.00} \\
         Voice + Action Interaction      & 2.64 & 2.36 & 2.42 & 2.33 & \underline{3.14} & \textbf{3.71} \\
         Object Interaction Sound        & 2.65 & 2.53 & 2.53 & \underline{2.82} & 2.65 & \textbf{3.82} \\
         Animal Sound                    & 2.21 & 2.64 & 2.77 & 2.36 & \underline{2.79} & \textbf{3.93} \\
         Ambient / Background Sound      & 2.67 & 2.89 & 2.89 & 2.38 & \underline{3.00} & \textbf{3.56} \\
         Special Effects (ASMR, etc.)    & 2.65 & 2.53 & 2.53 & 2.86 & \underline{3.18} & \textbf{3.53} \\
         Instruments \& Audio            & \underline{3.00} & 2.79 & 2.72 & 2.78 & 2.89 & \textbf{3.63} \\
         Dual-Channel Audio              & 2.86 & 3.00 & \underline{3.43} & 2.17 & 3.29 & \textbf{4.00} \\
         \bottomrule
         \end{tabular}%
         }
     \end{table}

%% file: tables/t2v/t2v_audio_adherence.tex
\begin{table}[ht!]
  \centering
  \caption{T2V detailed Audio Prompt Following evaluation results across fine-grained categories. Rating
  from 1 to 5, with higher scores indicating better performance.}
  \label{tab:t2v_audio_adherence_detailed}
  \fittowidth{%
  \begin{tabular}{lcccccc}
  \toprule
  \textbf{Category} & \textbf{Kling 2.6}~\cite{kling2p6} & \textbf{Kling 3.0}~\cite{kling3.0} & \textbf{Sora2 Pro}~\cite{sora2} & \textbf{Veo3.1}~\cite{veo3p1} & \textbf{Seedance 1.5}~\cite{seedance1p5} & \textbf{Seedance 2.0} \\
  \midrule
  Chinese Dialect / Accent        & 1.23 & \underline{1.86} & \underline{1.86} & 1.20 & 1.82 & \textbf{2.91} \\
  Chinese Multi-Person Dialogue   & 2.00 & \underline{2.79} & 2.21 & 1.80 & 2.36 & \textbf{3.29} \\
  Chinese Variety Show Voice      & 2.00 & 2.29 & \underline{2.71} & 1.57 & 2.43 & \textbf{3.29} \\
  Chinese Opera                   & 1.63 & 2.13 & \underline{2.33} & 1.29 & 1.75 & \textbf{3.50} \\
  English                         & 2.25 & 3.33 & \underline{3.64} & 3.00 & 3.25 & \textbf{4.25} \\
  Minority Languages              & 1.03 & 2.18 & 3.12 & 2.61 & \underline{3.26} & \textbf{3.74} \\
  Singing / Rap                   & 3.14 & 3.43 & \underline{3.67} & 2.17 & 2.14 & \textbf{3.71} \\
  Spatial Scene                   & 2.71 & \underline{2.86} & 2.71 & 2.50 & \underline{2.86} & \textbf{3.29} \\
  Off-Screen Voice                & 2.57 & \textbf{3.14} & \underline{3.00} & 1.83 & 2.71 & \textbf{3.14} \\
  Non-Verbal Voice                & 2.00 & 2.00 & 2.89 & 2.33 & \underline{3.00} & \textbf{3.67} \\
  Voice + Action Interaction      & 2.57 & 3.21 & \underline{3.50} & 2.33 & 3.21 & \textbf{3.86} \\
  Object Interaction Sound        & 2.41 & 2.00 & \underline{2.53} & \underline{2.53} & \underline{2.53} & \textbf{3.29} \\
  Animal Sound                    & 1.64 & 2.64 & \underline{2.77} & 2.21 & 2.50 & \textbf{3.86} \\
  Ambient / Background Sound      & 2.44 & 2.67 & \underline{3.22} & 2.38 & 2.78 & \textbf{3.89} \\
  Special Effects (ASMR, etc.)    & 2.18 & 2.35 & \underline{3.29} & 2.14 & 2.94 & \textbf{3.47} \\
  Instruments \& Audio            & 2.58 & 3.11 & \underline{3.61} & 3.06 & 2.79 & \textbf{3.89} \\
  Dual-Channel Audio              & 2.71 & 2.57 & \underline{2.86} & 2.17 & 2.43 & \textbf{3.29} \\
  \bottomrule
  \end{tabular}%
  }
\end{table}

%% file: tables/i2v/i2v_overall_score.tex
\begin{table}[H]
  \centering
  \caption{I2V overall evaluation results across video and audio dimensions (Rating from 1 to 5).}
  \label{tab:i2v_mos}
  \fittowidth{%
  \begin{tabular}{l | ccc | ccc}
  \toprule
  & \multicolumn{3}{c|}{\textbf{Video}} & \multicolumn{3}{c}{\textbf{Audio}} \\
  \cmidrule(lr){2-4}\cmidrule(lr){5-7}
  \multicolumn{1}{c|}{\multirow{-2}{*}{\textbf{Model}}} & \textbf{\makecell{Motion\\Quality}} & \textbf{\makecell{Video Prompt\\Following}} & \textbf{\makecell{Image\\Preservation}} & \textbf{\makecell{Quality
\&\\Expressiveness}} & \textbf{\makecell{Audio-Visual\\Sync}} & \textbf{\makecell{Audio Prompt\\Following}} \\
  \midrule
  Wan 2.6~\cite{wan2p6}          & 2.32 & 2.74 & 2.61 & 2.20 & 2.18 & 2.55 \\
  Kling 2.6~\cite{kling2p6}        & 2.52 & 2.55 & 2.98 & 2.21 & 2.27 & 2.21 \\
  Veo3.1~\cite{veo3p1}           & 2.65 & \underline{2.87} & 2.69 & 2.68 & 2.69 & 2.79 \\
  Seedance 1.5 Pro~\cite{seedance1p5} & 2.53 & 2.77 & 2.92 & \underline{3.07} & \underline{2.95} & \underline{3.10} \\
  Kling 3.0~\cite{kling3.0}        & \underline{2.80} & 2.78 & \underline{3.18} & 2.89 & 2.83 & 2.85 \\
  Seedance 2.0     & \textbf{3.35} & \textbf{3.46} & \textbf{3.31} & \textbf{3.61} & \textbf{3.54} & \textbf{3.70} \\
  \bottomrule
  \end{tabular}%
  }
\end{table}

%% file: tables/i2v/i2v_overall_analysis.tex
\begin{table}[ht!]
  \centering
  \caption{I2V usability and satisfaction rates across video and audio dimensions.}
  \vspace{-0.1cm}
  \label{tab:i2v_rates}
  \fittowidth{%
  \begin{tabular}{l | ccc | ccc}
  \toprule
  & \multicolumn{3}{c|}{\textbf{Video}} & \multicolumn{3}{c}{\textbf{Audio}} \\
  \cmidrule(lr){2-4}\cmidrule(lr){5-7}
  \multicolumn{1}{c|}{\multirow{-2}{*}{\textbf{Model}}} & \textbf{\makecell{Motion\\Quality}} & \textbf{\makecell{Video Prompt\\Following}} & \textbf{\makecell{Image\\Preservation}} & \textbf{\makecell{Quality
\&\\Expressiveness}} & \textbf{\makecell{Audio-Visual\\Sync}} & \textbf{\makecell{Audio Prompt\\Following}} \\
  \midrule
  \multicolumn{7}{c}{\textbf{\textit{Usability Rate (score $\geq$ 3)}}} \\
  \midrule
  Wan 2.6~\cite{wan2p6}          & 32.71\% & 59.02\% & 59.77\% & 27.03\% & 25.68\% & 51.80\% \\
  Kling 2.6~\cite{kling2p6}        & 48.72\% & 48.72\% & 80.59\% & 27.19\% & 31.14\% & 30.26\% \\
  Veo3.1~\cite{veo3p1}           & 59.69\% & \underline{65.50\%} & 67.05\% & 60.75\% & 59.35\% & 56.54\% \\
  Seedance 1.5 Pro~\cite{seedance1p5} & 51.44\% & 58.99\% & 80.58\% & \underline{93.99\%} & \underline{80.26\%} & \underline{70.39\%} \\
  Kling 3.0~\cite{kling3.0}        & \underline{68.00\%} & 60.00\% & \underline{90.55\%} & 77.39\% & 68.70\% & 66.52\% \\
  Seedance 2.0     & \textbf{87.05\%} & \textbf{88.85\%} & \textbf{91.37\%} & \textbf{97.42\%} & \textbf{91.85\%} & \textbf{92.27\%} \\
  \midrule
  \multicolumn{7}{c}{\textbf{\textit{Satisfaction Rate (score $\geq$ 4)}}} \\
  \midrule
  Wan 2.6~\cite{wan2p6}          & 2.63\% & 16.54\% & 7.52\% & 0.45\% & 1.80\% & 9.91\% \\
  Kling 2.6~\cite{kling2p6}        & 5.86\% & 9.52\% & 19.05\% & 1.32\% & 2.19\% & 5.70\% \\
  Veo3.1~\cite{veo3p1}           & 7.36\% & \underline{20.54\%} & 6.20\% & 7.48\% & 10.28\% & 24.30\% \\
  Seedance 1.5 Pro~\cite{seedance1p5} & 3.60\% & 18.71\% & 12.59\% & \underline{13.30\%} & \underline{15.45\%} & \underline{37.77\%} \\
  Kling 3.0~\cite{kling3.0}        & \underline{12.00\%} & 18.91\% & \underline{27.27\%} & 12.61\% & 15.22\% & 20.87\% \\
  Seedance 2.0     & \textbf{43.88\%} & \textbf{47.48\%} & \textbf{38.85\%} & \textbf{57.08\%} & \textbf{54.94\%} & \textbf{63.52\%} \\
  \bottomrule
  \end{tabular}%
  }
\end{table}

%% file: tables/i2v/i2v_analysis_visual.tex
\begin{table}[H]
\centering
\caption{Fine-grained I2V visual evaluation on prompt abstraction tasks. Rating from 1 to 5. MQ means motion quality, IP means image preservation, and VPF means video prompt following.}
\label{tab:i2v_vis_prompt}
\fittowidth{%
\begin{tabular}{l cccccc}
\toprule
\multicolumn{1}{c}{\multirow{3}{*}{\textbf{Model}}} & \multicolumn{3}{c}{\textbf{UGC Creative / Portrait}} & \multicolumn{3}{c}{\textbf{Script-Controlled (15s)}} \\
\cmidrule(lr){2-4}\cmidrule(lr){5-7}
& \textbf{MQ} & \textbf{IP} & \textbf{VPF} & \textbf{MQ} & \textbf{IP} & \textbf{VPF} \\
\midrule
Kling 2.6~\cite{kling2p6}        & 2.33 & 3.00 & 2.33 & 2.47 & 2.73 & 2.13 \\
Wan 2.6~\cite{wan2p6}          & 2.60 & 3.00 & 3.13 & 2.14 & 2.36 & 2.50 \\
Veo 3.1~\cite{veo3p1}          & 2.53 & 2.87 & \textbf{3.87} & 2.58 & 2.33 & 2.33 \\
Seedance 1.5 Pro~\cite{seedance1p5} & 2.87 & 3.07 & 3.13 & 2.80 & \underline{2.87} & 2.53 \\
Kling 3.0~\cite{kling3.0}        & \underline{3.07} & \underline{3.27} & 2.73 & \underline{3.07} & \underline{2.87} & \underline{3.00} \\
Seedance 2.0     & \textbf{3.40} & \textbf{3.40} & \underline{3.53} & \textbf{3.40} & \textbf{3.13} & \textbf{3.87} \\
\bottomrule
\end{tabular}%
}
\end{table}

\paragraph{Prompt Abstraction.} This category tests UGC-style creative prompts and script-controlled generation. Seedance 2.0 leads on all three metrics for script-controlled (15s) generation and on MQ and IP for UGC creative/portrait (Table~\ref{tab:i2v_vis_prompt}). The gap is largest on script-controlled (15s) generation: VPF 3.87 vs.\ Kling 3.0's 3.00 and Wan 2.6's 2.50. Veo 3.1 scores 3.87 on VPF for UGC creative/portrait---the best in that sub-category---but its MQ (2.53) and IP (2.87) are much lower. Kling 2.6 falls below 2.5 on MQ for both sub-categories, and Wan 2.6 drops to 2.14 on script-controlled MQ. Open-ended and script-based prompts remain difficult for most models; only Seedance 2.0 and Kling 3.0 consistently exceed 3.0 on MQ.

\begin{table}[H]
  \centering
  \caption{Fine-grained I2V visual evaluation on complex instruction following. Rating from 1 to 5.}
  \label{tab:i2v_vis_instr}
  \fittowidth{%
  \begin{tabular}{l ccccccccc}
  \toprule
  \multicolumn{1}{c}{\multirow{2}{*}{\rule{0pt}{3ex}\textbf{Model}}} & \multicolumn{3}{c}{\textbf{\makecell{New Entity\\(Size Rel.)}}} & \multicolumn{3}{c}{\textbf{\makecell{Compound\\Multi-Instr.}}} & \multicolumn{3}{c}{\textbf{\makecell{Degree\\Adverbs}}} \\
  \cmidrule(lr){2-4}\cmidrule(lr){5-7}\cmidrule(lr){8-10}
  & \textbf{MQ} & \textbf{IP} & \textbf{VPF} & \textbf{MQ} & \textbf{IP} & \textbf{VPF} & \textbf{MQ} & \textbf{IP} & \textbf{VPF} \\
  \midrule
  Kling 2.6~\cite{kling2p6}        & 2.50 & 2.88 & 3.13 & 2.38 & 2.88 & 2.00 & 2.47 & 2.87 & 2.33 \\
  Wan 2.6~\cite{wan2p6}          & 2.63 & 2.63 & 2.75 & 2.29 & 2.43 & \underline{2.57} & 2.38 & 3.00 & 2.77 \\
  Veo 3.1~\cite{veo3p1}          & 2.50 & 2.88 & \underline{3.38} & 2.57 & 2.29 & 2.29 & 2.79 & 2.64 & 2.79 \\
  Seedance 1.5 Pro~\cite{seedance1p5} & \underline{3.25} & 2.75 & 2.63 & 2.13 & 2.63 & 2.25 & \underline{2.80} & 3.00 & \underline{3.07} \\
  Kling 3.0~\cite{kling3.0}        & 2.75 & \underline{3.00} & 3.00 & \underline{2.88} & \underline{3.25} & 2.50 & 2.60 & \underline{3.07} & 2.67 \\
  Seedance 2.0     & \textbf{3.75} & \textbf{3.25} & \textbf{3.88} & \textbf{4.00} & \textbf{3.38} & \textbf{3.75} & \textbf{3.20} & \textbf{3.40} & \textbf{3.40} \\
  \bottomrule
  \end{tabular}%
  }
\end{table}

\paragraph{Complex Instructions.} Compound multi-instruction is where Seedance 2.0 pulls furthest ahead: MQ 4.00 and VPF 3.75, outscoring Kling 3.0 by over 1.0 point on MQ (Table~\ref{tab:i2v_vis_instr}). Seedance 1.5 Pro scored 2.13/2.25 on MQ/VPF for this sub-category, so complex instruction handling improved by nearly 2 points in the 2.0 generation. Veo 3.1 reaches 3.38 on VPF for new entity tasks---close to Seedance 2.0's 3.88---but its MQ (2.50) lags, meaning it follows the instruction but produces weaker motion. Degree adverbs is the tightest sub-category: Seedance 2.0 scores 3.20/3.40/3.40 with Seedance 1.5 Pro not far behind at 2.80/3.00/3.07.

\begin{table}[H]
\centering
\caption{Fine-grained I2V visual evaluation on complex camera work. Rating from 1 to 5.}
\label{tab:i2v_vis_camera}
\fittowidth{%
\begin{tabular}{l ccccccccc}
\toprule
\multicolumn{1}{c}{\multirow{3}{*}{\textbf{Model}}} & \multicolumn{3}{c}{\textbf{\makecell{Combined\\Shot Instr.}}} & \multicolumn{3}{c}{\textbf{\makecell{Adv. Camera\\Movement}}} & \multicolumn{3}{c}{\textbf{\makecell{Difficult Shots \&\\Special Tech.}}} \\
\cmidrule(lr){2-4}\cmidrule(lr){5-7}\cmidrule(lr){8-10}
& \textbf{MQ} & \textbf{IP} & \textbf{VPF} & \textbf{MQ} & \textbf{IP} & \textbf{VPF} & \textbf{MQ} & \textbf{IP} & \textbf{VPF} \\
\midrule
Kling 2.6~\cite{kling2p6}        & 2.43 & 2.93 & 2.43 & 2.29 & 2.57 & \underline{2.86} & 2.50 & \underline{3.25} & \textbf{3.38} \\
Wan 2.6~\cite{wan2p6}          & 2.33 & 2.93 & 2.93 & 2.14 & 2.14 & 2.71 & 2.25 & 2.50 & \underline{3.00} \\
Veo 3.1~\cite{veo3p1}          & 2.80 & 3.00 & 3.00 & \underline{2.67} & 2.50 & 2.33 & \underline{2.75} & 2.38 & 2.75 \\
Seedance 1.5 Pro~\cite{seedance1p5} & 2.67 & \underline{3.07} & \underline{3.07} & 2.43 & \underline{2.71} & 2.57 & 2.50 & 2.63 & \underline{3.00} \\
Kling 3.0~\cite{kling3.0}        & \underline{3.20} & \textbf{3.27} & 2.93 & \textbf{2.71} & \underline{2.71} & 2.43 & 2.63 & \underline{3.25} & 2.38 \\
Seedance 2.0     & \textbf{3.47} & \textbf{3.27} & \textbf{3.53} & \textbf{2.71} & \textbf{3.00} & \textbf{3.14} & \textbf{3.50} & \textbf{3.88} & \underline{3.00} \\
\bottomrule
\end{tabular}%
}
\end{table}

\paragraph{Complex Camera.} Seedance 2.0 leads on MQ for all three camera sub-categories and on VPF for combined shot instructions and advanced camera movement (Table~\ref{tab:i2v_vis_camera}). On difficult shots \& special techniques, it scores 3.88 on IP---the highest single value in this table---while Kling 3.0 reaches 3.25. Kling 2.6 scores 3.38 on VPF for difficult shots---the best in that sub-category---handling special techniques better than general camera work. Advanced camera movement is the hardest sub-category: Seedance 2.0 and Kling 3.0 tie on MQ (2.71), and no model exceeds 3.14 on any metric. Camera flexibility is an area where all models have room to improve; advanced movement scores stay below 3.2 across the board.

\begin{table}[ht!]
\centering
\caption{Fine-grained I2V visual evaluation on complex motion. Rating from 1 to 5.}
\vspace{-0.1cm}
\label{tab:i2v_vis_motion}
\fittowidth{%
\begin{tabular}{l cccccccccccc}
\toprule
\multicolumn{1}{c}{\multirow{3}{*}{\textbf{Model}}} & \multicolumn{3}{c}{\textbf{Sports}} & \multicolumn{3}{c}{\textbf{Fine Motion}} & \multicolumn{3}{c}{\textbf{Micro-Expr. \& Emotion}} & \multicolumn{3}{c}{\textbf{Combat Visual Effects}} \\
\cmidrule(lr){2-4}\cmidrule(lr){5-7}\cmidrule(lr){8-10}\cmidrule(lr){11-13}
& \textbf{MQ} & \textbf{IP} & \textbf{VPF} & \textbf{MQ} & \textbf{IP} & \textbf{VPF} & \textbf{MQ} & \textbf{IP} & \textbf{VPF} & \textbf{MQ} & \textbf{IP} & \textbf{VPF} \\
\midrule
Kling 2.6~\cite{kling2p6}        & \underline{3.00} & 3.07 & 2.53 & 2.40 & \underline{3.27} & 2.47 & 2.88 & 3.38 & 2.63 & \underline{2.63} & 2.75 & 2.50 \\
Wan 2.6~\cite{wan2p6}          & 2.00 & 2.64 & 2.29 & 2.57 & 2.86 & \underline{3.00} & 2.38 & 2.38 & \underline{3.13} & 2.14 & 1.86 & 2.14 \\
Veo 3.1~\cite{veo3p1}          & 2.54 & 2.92 & 2.69 & 2.57 & 3.00 & 2.79 & 3.00 & 3.00 & 3.00 & 2.50 & 2.75 & \underline{2.88} \\
Seedance 1.5 Pro~\cite{seedance1p5} & 2.40 & 3.07 & 2.53 & 2.33 & 3.20 & 2.73 & 2.88 & 3.13 & \underline{3.13} & 2.25 & 2.63 & 2.50 \\
Kling 3.0~\cite{kling3.0}        & 2.71 & \underline{3.29} & \underline{3.00} & \underline{2.80} & \textbf{3.47} & \underline{3.00} & \underline{3.13} & \underline{3.50} & 2.88 & 2.25 & \underline{2.88} & 2.13 \\
Seedance 2.0     & \textbf{3.73} & \textbf{3.47} & \textbf{3.93} & \textbf{3.33} & \textbf{3.47} & \textbf{3.53} & \textbf{3.63} & \textbf{3.63} & \textbf{4.00} & \textbf{3.63} & \textbf{3.25} & \textbf{3.13} \\
\bottomrule
\end{tabular}%
}
\end{table}

\paragraph{Complex Motion.} Seedance 2.0's strongest results here are sports (MQ 3.73, VPF 3.93) and micro-expression \& emotion (VPF 4.00), as shown in Table~\ref{tab:i2v_vis_motion}. Combat visual effects shows the widest gap: Seedance 2.0 scores MQ 3.63 vs.\ Kling 3.0's 2.25 and Seedance 1.5 Pro's 2.25---a 1.38-point difference. Expression and gaze vividness improved substantially over Seedance 1.5 Pro (micro-expression MQ: 2.88 $\to$ 3.63). Kling 3.0 is competitive on fine motion IP (3.47, tying with Seedance 2.0) and micro-expression IP (3.50), preserving image identity well even when its motion quality lags. Wan 2.6 scores 1.86 on combat IP---the lowest value across all visual tables.

\begin{table}[ht!]
\centering
\caption{Fine-grained I2V visual evaluation on complex interaction. Rating from 1 to 5.}
\vspace{-0.1cm}
\label{tab:i2v_vis_interaction}
\fittowidth{%
\begin{tabular}{l ccccccccc}
\toprule
\multicolumn{1}{c}{\multirow{3}{*}{\textbf{Model}}} & \multicolumn{3}{c}{\textbf{Group Motion}} & \multicolumn{3}{c}{\textbf{Same-Type Interaction}} & \multicolumn{3}{c}{\textbf{Cross-Type Interaction}} \\
\cmidrule(lr){2-4}\cmidrule(lr){5-7}\cmidrule(lr){8-10}
& \textbf{MQ} & \textbf{IP} & \textbf{VPF} & \textbf{MQ} & \textbf{IP} & \textbf{VPF} & \textbf{MQ} & \textbf{IP} & \textbf{VPF} \\
\midrule
Kling 2.6~\cite{kling2p6}        & 2.38 & 2.50 & \underline{2.75} & \underline{2.91} & \underline{3.36} & 2.45 & 2.88 & \underline{3.13} & 2.75 \\
Wan 2.6~\cite{wan2p6}          & 2.13 & 2.25 & 2.50 & 2.45 & 2.82 & 2.91 & 2.38 & \underline{3.13} & 2.88 \\
Veo 3.1~\cite{veo3p1}          & \underline{2.63} & 2.50 & \underline{2.75} & 2.40 & 2.50 & 2.90 & 3.00 & 2.88 & 3.00 \\
Seedance 1.5 Pro~\cite{seedance1p5} & 2.50 & \underline{2.63} & \textbf{2.88} & 2.45 & 2.91 & \underline{3.18} & 2.88 & \textbf{3.25} & 3.13 \\
Kling 3.0~\cite{kling3.0}        & 2.50 & \textbf{3.00} & 2.50 & 2.73 & 3.27 & 2.55 & \underline{3.38} & \underline{3.13} & \underline{3.63} \\
Seedance 2.0     & \textbf{3.00} & \textbf{3.00} & \textbf{2.88} & \textbf{3.64} & \textbf{3.82} & \textbf{3.91} & \textbf{3.50} & \textbf{3.25} & \textbf{4.00} \\
\bottomrule
\end{tabular}%
}
\end{table}

\paragraph{Complex Interaction.} Same-type interaction (MQ 3.64, IP 3.82, VPF 3.91) and cross-type interaction (VPF 4.00) are Seedance 2.0's strongest results in Table~\ref{tab:i2v_vis_interaction}. Group motion is hard for everyone---Seedance 2.0 scores 3.00/3.00/2.88, and most competitors hover near 2.5. Kling 3.0 scores 3.63 on cross-type VPF, close to Seedance 2.0's 4.00, and 3.38 on MQ, handling inter-species or human-object interactions reasonably well. Kling 2.6 scores 3.36 on same-type IP but lags on MQ (2.91) and VPF (2.45).

\begin{table}[ht!]
    \centering
    \caption{Fine-grained I2V visual evaluation on creative generation. Rating from 1 to 5.}
    \vspace{-0.1cm}
    \label{tab:i2v_vis_creative}
    \fittowidth{%
    \begin{tabular}{l cccccccccccc}
    \toprule
    \multicolumn{1}{c}{\multirow{2}{*}{\rule{0pt}{3ex}\textbf{Model}}} & \multicolumn{3}{c}{\textbf{\makecell{Counter-\\Reality}}} & \multicolumn{3}{c}{\textbf{\makecell{Design\\Instructions}}} & \multicolumn{3}{c}{\textbf{\makecell{Visual Effects\\(Transformation)}}} & \multicolumn{3}{c}{\textbf{Holidays}} \\
    \cmidrule(lr){2-4}\cmidrule(lr){5-7}\cmidrule(lr){8-10}\cmidrule(lr){11-13}
    & \textbf{MQ} & \textbf{IP} & \textbf{VPF} & \textbf{MQ} & \textbf{IP} & \textbf{VPF} & \textbf{MQ} & \textbf{IP} & \textbf{VPF} & \textbf{MQ} & \textbf{IP} & \textbf{VPF} \\
    \midrule
    Kling 2.6~\cite{kling2p6}        & 2.36 & 2.71 & 2.57 & 2.20 & \underline{2.80} & 2.20 & 2.43 & 2.57 & 2.71 & 2.43 & 2.86 & 2.71 \\
    Wan 2.6~\cite{wan2p6}          & 2.08 & 2.08 & 2.54 & 2.47 & 2.60 & 2.53 & 2.13 & 2.50 & 2.75 & 2.00 & 2.25 & 2.75 \\
    Veo 3.1~\cite{veo3p1}          & \underline{2.43} & 2.57 & \underline{2.86} & \underline{2.67} & 2.60 & \textbf{2.93} & \underline{3.00} & 2.50 & \underline{3.38} & 2.50 & 2.00 & 2.38 \\
    Seedance 1.5 Pro~\cite{seedance1p5} & 2.27 & \underline{2.80} & 2.40 & 2.33 & 2.60 & 2.20 & 2.75 & 2.88 & 3.25 & 2.38 & \underline{2.88} & 2.38 \\
    Kling 3.0~\cite{kling3.0}        & 2.27 & \textbf{3.07} & 2.47 & 2.60 & \textbf{2.87} & 2.33 & 2.63 & \underline{3.13} & 2.50 & \underline{2.71} & 2.86 & \underline{3.00} \\
    Seedance 2.0     & \textbf{2.93} & \textbf{3.07} & \textbf{3.07} & \textbf{3.13} & \textbf{2.87} & \underline{2.87} & \textbf{3.50} & \textbf{3.38} & \textbf{3.50} & \textbf{3.00} & \textbf{3.00} & \textbf{3.38} \\
    \bottomrule
    \end{tabular}%
    }
\end{table}

\paragraph{Creative.} Seedance 2.0 leads on MQ for all four creative sub-categories (Table~\ref{tab:i2v_vis_creative}), with visual effects (transformation) as its best at 3.50/3.38/3.50. Realistic and 3D visual effects render fluidly, and the model preserves special art styles (felt, oil painting, Chinese gongbi) while matching motion to the style. Veo 3.1 is close on VPF for visual effects (3.38) and design instructions (2.93), but its MQ trails. Holidays is weak for most models---Veo 3.1 drops to 2.00 on IP, Wan 2.6 to 2.00/2.25. No model besides Seedance 2.0 exceeds 3.50 on any metric in this category.

\begin{table}[H]
\centering
\caption{Fine-grained I2V visual evaluation on physical laws. Rating from 1 to 5.}
\vspace{-0.1cm}
\label{tab:i2v_vis_physics}
\fittowidth{%
\begin{tabular}{l ccccccccc}
\toprule
\multicolumn{1}{c}{\multirow{3}{*}{\textbf{Model}}} & \multicolumn{3}{c}{\textbf{Natural Phenomena}} & \multicolumn{3}{c}{\textbf{Physical Phen. (Prof.)}} & \multicolumn{3}{c}{\textbf{Physical Feedback (Daily)}} \\
\cmidrule(lr){2-4}\cmidrule(lr){5-7}\cmidrule(lr){8-10}
& \textbf{MQ} & \textbf{IP} & \textbf{VPF} & \textbf{MQ} & \textbf{IP} & \textbf{VPF} & \textbf{MQ} & \textbf{IP} & \textbf{VPF} \\
\midrule
Kling 2.6~\cite{kling2p6}        & 2.67 & 3.11 & 2.67 & 2.57 & \textbf{3.50} & 2.50 & 2.29 & 2.86 & 2.50 \\
Wan 2.6~\cite{wan2p6}          & 2.38 & 2.00 & 2.88 & 2.38 & 2.77 & 2.54 & 2.27 & 2.80 & 2.67 \\
Veo 3.1~\cite{veo3p1}          & 2.67 & 2.56 & 2.78 & \underline{2.79} & 3.00 & \underline{3.07} & 2.46 & 2.92 & 2.69 \\
Seedance 1.5 Pro~\cite{seedance1p5} & 2.33 & 2.89 & 2.56 & 2.14 & 3.07 & 2.36 & 2.13 & 2.93 & \underline{2.80} \\
Kling 3.0~\cite{kling3.0}        & \underline{3.00} & \textbf{3.56} & \underline{3.00} & 2.64 & \underline{3.36} & 2.93 & \underline{2.73} & \textbf{3.13} & 2.73 \\
Seedance 2.0     & \textbf{3.33} & \underline{3.44} & \textbf{3.33} & \textbf{3.14} & \underline{3.36} & \textbf{3.36} & \textbf{2.87} & \underline{3.07} & \textbf{2.93} \\
\bottomrule
\end{tabular}%
}
\end{table}

\paragraph{Physical Laws.} Seedance 2.0 leads on MQ across all three physical laws sub-categories (Table~\ref{tab:i2v_vis_physics}), scoring 2.87--3.33. Kling 3.0 outscores Seedance 2.0 on IP for natural phenomena (3.56 vs.\ 3.44) and ties on professional phenomena (3.36)---one of the few areas where a competitor beats Seedance 2.0 on a specific metric. Kling 2.6 scores 3.50 on professional phenomena IP, its highest value in any visual table, though its MQ (2.57) and VPF (2.50) stay low. Physical laws is difficult across the board: Seedance 1.5 Pro scores below 2.4 on MQ for all three sub-categories, and motion stability during physics simulations remains a challenge for every model.

\begin{table}[H]
\centering
\caption{Fine-grained I2V visual evaluation on complex reference images. Rating from 1 to 5.}
\label{tab:i2v_vis_reference}
\fittowidth{%
\begin{tabular}{l cccccc}
\toprule
\multicolumn{1}{c}{\multirow{2}{*}{\rule{0pt}{3ex}\textbf{Model}}} & \multicolumn{3}{c}{\textbf{\makecell{High Information\\Density}}} & \multicolumn{3}{c}{\textbf{\makecell{Multi-Ethnicity /\\Skin Tone}}} \\
\cmidrule(lr){2-4}\cmidrule(lr){5-7}
& \textbf{MQ} & \textbf{IP} & \textbf{VPF} & \textbf{MQ} & \textbf{IP} & \textbf{VPF} \\
\midrule
Kling 2.6~\cite{kling2p6}        & 2.67 & 3.07 & 3.00 & 2.60 & 3.07 & 2.73 \\
Wan 2.6~\cite{wan2p6}          & 2.20 & 2.53 & 3.00 & 2.57 & 3.00 & 2.71 \\
Veo 3.1~\cite{veo3p1}          & 2.67 & 2.80 & 2.87 & 2.82 & 2.45 & 2.64 \\
Seedance 1.5 Pro~\cite{seedance1p5} & 2.67 & 3.00 & \underline{3.13} & 2.73 & 3.00 & \underline{3.00} \\
Kling 3.0~\cite{kling3.0}        & \underline{2.87} & \underline{3.33} & \underline{3.13} & \underline{3.21} & \textbf{3.43} & 2.93 \\
Seedance 2.0     & \textbf{3.40} & \textbf{3.40} & \textbf{3.73} & \textbf{3.47} & \underline{3.33} & \textbf{3.53} \\
\bottomrule
\end{tabular}%
}
\end{table}

\paragraph{Complex Reference.} Seedance 2.0 leads on MQ and VPF for both complex reference sub-categories, and on IP for high information density (Table~\ref{tab:i2v_vis_reference}). High information density VPF (3.73) outscores Kling 3.0 (3.13) by 0.60 points. Kling 3.0 outscores Seedance 2.0 on multi-ethnicity IP (3.43 vs.\ 3.33), one of the few metrics where a competitor takes the lead. The visual average confirms the overall ranking: Seedance 2.0 leads on MQ (3.35), IP (3.31), and VPF (3.46). Kling 3.0 is second on IP (3.18) and third on VPF (2.78), ahead of Seedance 1.5 Pro (2.77). Wan 2.6 ranks last on MQ (2.32) and IP (2.61).

%% file: tables/i2v/i2v_analysis_audio.tex
\begin{table}[ht!]
\centering
\caption{Fine-grained I2V audio evaluation on Chinese voice. AQ = Audio Quality \& Expressiveness, AVS = Audio-Visual Sync, APF = Audio Prompt Following. Rating from 1 to 5.}
\vspace{-0.1cm}
\label{tab:i2v_aud_chinese}
\fittowidth{%
\begin{tabular}{l cccccccccccc}
\toprule
\multicolumn{1}{c}{\multirow{2}{*}{\rule{0pt}{3ex}\textbf{Model}}} & \multicolumn{3}{c}{\textbf{\makecell{Chinese Dialect /\\Lip Sync}}} & \multicolumn{3}{c}{\textbf{\makecell{Chinese\\Conversation}}} & \multicolumn{3}{c}{\textbf{\makecell{Variety Show\\Voice}}} & \multicolumn{3}{c}{\textbf{\makecell{Chinese\\Opera}}} \\
\cmidrule(lr){2-4}\cmidrule(lr){5-7}\cmidrule(lr){8-10}\cmidrule(lr){11-13}
& \textbf{AQ} & \textbf{AVS} & \textbf{APF} & \textbf{AQ} & \textbf{AVS} & \textbf{APF} & \textbf{AQ} & \textbf{AVS} & \textbf{APF} & \textbf{AQ} & \textbf{AVS} & \textbf{APF} \\
\midrule
Kling 2.6~\cite{kling2p6}        & 2.00 & 2.25 & 2.25 & 2.33 & 2.08 & 2.33 & 2.13 & 2.13 & 2.25 & 2.50 & 2.38 & 2.13 \\
Wan 2.6~\cite{wan2p6}          & 2.46 & 2.15 & 2.92 & 2.33 & 2.08 & 2.75 & 2.43 & \underline{3.00} & 3.00 & 2.38 & 2.38 & 2.13 \\
Veo 3.1~\cite{veo3p1}          & 2.09 & 2.45 & 2.18 & 2.20 & 2.50 & 2.10 & 2.14 & 2.29 & 1.86 & 2.25 & 2.75 & 1.75 \\
Seedance 1.5 Pro~\cite{seedance1p5} & 2.92 & 3.00 & 2.92 & 2.25 & \underline{2.83} & \underline{3.08} & \underline{3.00} & 2.75 & \underline{3.13} & \underline{3.00} & 2.88 & \underline{2.38} \\
Kling 3.0~\cite{kling3.0}        & \underline{3.17} & \underline{3.08} & \underline{3.08} & \underline{3.09} & 2.73 & 3.00 & 2.75 & 2.88 & \underline{3.13} & 2.88 & \underline{3.00} & 1.88 \\
Seedance 2.0     & \textbf{3.23} & \textbf{3.46} & \textbf{3.31} & \textbf{3.92} & \textbf{3.42} & \textbf{4.08} & \textbf{3.13} & \textbf{4.00} & \textbf{4.00} & \textbf{3.75} & \textbf{3.38} & \textbf{2.50} \\
\bottomrule
\end{tabular}%
}
\end{table}

\paragraph{Chinese Voice.} Seedance 2.0 leads on AQ for all four Chinese voice sub-categories in Table~\ref{tab:i2v_aud_chinese}, scoring 3.13--3.92. Chinese dialogue carries emotional nuance, and common dialects and accents come through clearly. Chinese conversation is its strongest (AQ 3.92, APF 4.08), with a 0.83-point AQ lead over Kling 3.0 and a 1.67-point lead over Seedance 1.5 Pro. Variety show voice reaches 4.00 on both AVS and APF, the highest sync score in this table. Chinese opera is a weak spot for all models on prompt following---Seedance 2.0 scores only 2.50 on APF, though its AQ (3.75) is best. Kling 3.0 is the closest competitor on dialect/lip sync (AQ 3.17 vs.\ 3.23) and opera sync (AVS 3.00 vs.\ 3.38). Veo 3.1, Wan 2.6, and Kling 2.6 score below 2.5 on AQ for most sub-categories, with Kling 2.6 at 2.00 on dialect---the lowest AQ in this table.

\begin{table}[ht!]
\centering
\caption{Fine-grained I2V audio evaluation on non-Chinese voice. Rating from 1 to 5.}
\vspace{-0.1cm}
\label{tab:i2v_aud_nonchi}
\fittowidth{%
\begin{tabular}{l cccccccccccccccccc}
\toprule
& \multicolumn{3}{c}{\textbf{English}} & \multicolumn{3}{c}{\textbf{Japanese}} & \multicolumn{3}{c}{\textbf{Korean}} & \multicolumn{3}{c}{\textbf{Indonesian}} & \multicolumn{3}{c}{\textbf{Portuguese}} & \multicolumn{3}{c}{\textbf{Spanish}} \\
\cmidrule(lr){2-4}\cmidrule(lr){5-7}\cmidrule(lr){8-10}\cmidrule(lr){11-13}\cmidrule(lr){14-16}\cmidrule(lr){17-19}
\multicolumn{1}{c}{\multirow{-2}{*}{\textbf{Model}}} & \textbf{AQ} & \textbf{AVS} & \textbf{APF} & \textbf{AQ} & \textbf{AVS} & \textbf{APF} & \textbf{AQ} & \textbf{AVS} & \textbf{APF} & \textbf{AQ} & \textbf{AVS} & \textbf{APF} & \textbf{AQ} & \textbf{AVS} & \textbf{APF} & \textbf{AQ} & \textbf{AVS} & \textbf{APF} \\
\midrule
Kling 2.6~\cite{kling2p6}        & 2.40 & 2.33 & 2.40 & 1.88 & 2.25 & 1.00 & 1.57 & 2.29 & 1.00 & 2.29 & 2.00 & 1.29 & 2.38 & 2.38 & 1.38 & 2.29 & 2.29 & 1.57 \\
Wan 2.6~\cite{wan2p6}          & 2.33 & 2.40 & 2.40 & 2.25 & 2.25 & 2.13 & 2.13 & 2.25 & 2.75 & 2.29 & 1.71 & 1.71 & 2.13 & 2.13 & 2.38 & 2.20 & 2.20 & 2.20 \\
Veo 3.1~\cite{veo3p1}          & 3.21 & 2.71 & 2.93 & 3.00 & 2.57 & 2.71 & 2.88 & 2.75 & 3.13 & 2.83 & 2.33 & \underline{3.00} & 2.86 & 2.86 & 3.00 & 3.00 & 3.00 & \underline{3.83} \\
Seedance 1.5 Pro~\cite{seedance1p5} & \underline{3.33} & \underline{3.13} & \underline{3.40} & \underline{3.13} & \underline{3.13} & \textbf{3.63} & \underline{3.13} & \underline{3.00} & \textbf{3.50} & \underline{3.14} & 3.14 & \underline{3.00} & \underline{2.88} & \underline{3.13} & \textbf{4.00} & \underline{3.29} & \underline{3.14} & 3.43 \\
Kling 3.0~\cite{kling3.0}        & 3.07 & 3.00 & 2.80 & 2.88 & 2.88 & 2.50 & 2.63 & 2.88 & 3.25 & 3.00 & \underline{3.57} & 1.71 & \underline{2.88} & 2.88 & 3.00 & 3.00 & 3.00 & 3.00 \\
Seedance 2.0     & \textbf{4.00} & \textbf{3.93} & \textbf{4.20} & \textbf{4.00} & \textbf{3.63} & \underline{3.13} & \textbf{3.75} & \textbf{3.38} & \underline{3.38} & \textbf{3.71} & \textbf{3.71} & \textbf{4.14} & \textbf{3.50} & \textbf{3.63} & \underline{3.63} & \textbf{4.14} & \textbf{4.14} & \textbf{4.00} \\
\bottomrule
\end{tabular}%
}
\end{table}

\paragraph{Non-Chinese Voice.} Seedance 2.0 scores at least 3.50 on AQ for all six non-Chinese languages in Table~\ref{tab:i2v_aud_nonchi}, peaking on Spanish (AQ 4.14, AVS 4.14) and English (AQ 4.00, APF 4.20). English prompt following reaches 4.20---the highest APF in this table. Indonesian APF (4.14) is also strong, providing the second-highest prompt following score. Seedance 1.5 Pro is second overall, with competitive scores on Portuguese APF (4.00) and Japanese APF (3.63), occasionally matching or exceeding Seedance 2.0 on prompt following. Veo 3.1 scores 3.83 on Spanish APF, its single best result across all audio tables, but its AQ and AVS stay around 2.7--3.0. Wan 2.6 and Kling 2.6 fall far behind: Kling 2.6 scores 1.00 on APF for both Japanese and Korean, and Wan 2.6 drops to 1.71 on Indonesian AVS and APF.

\begin{table}[ht!]
\centering
\caption{Fine-grained I2V audio evaluation on composite voice tasks. Rating from 1 to 5.}
\vspace{-0.1cm}
\label{tab:i2v_aud_composite}
\fittowidth{%
\begin{tabular}{l cccccccccccc}
\toprule
\multicolumn{1}{c}{\multirow{2}{*}{\rule{0pt}{3ex}\textbf{Model}}} & \multicolumn{3}{c}{\textbf{\makecell{Singing /\\Rap}}} & \multicolumn{3}{c}{\textbf{\makecell{Off-Screen\\Voice}}} & \multicolumn{3}{c}{\textbf{\makecell{Spatial\\Scene}}} & \multicolumn{3}{c}{\textbf{\makecell{Non-Verbal\\Voice}}} \\
\cmidrule(lr){2-4}\cmidrule(lr){5-7}\cmidrule(lr){8-10}\cmidrule(lr){11-13}
& \textbf{AQ} & \textbf{AVS} & \textbf{APF} & \textbf{AQ} & \textbf{AVS} & \textbf{APF} & \textbf{AQ} & \textbf{AVS} & \textbf{APF} & \textbf{AQ} & \textbf{AVS} & \textbf{APF} \\
\midrule
Kling 2.6~\cite{kling2p6}        & 2.67 & 2.33 & 3.22 & 2.50 & 2.38 & 2.50 & 2.20 & 2.20 & 2.40 & 2.23 & 2.38 & 2.46 \\
Wan 2.6~\cite{wan2p6}          & 2.50 & 2.30 & 3.00 & 2.43 & 2.14 & 2.57 & 2.20 & 2.30 & 2.40 & 2.08 & 2.15 & 2.54 \\
Veo 3.1~\cite{veo3p1}          & \underline{3.30} & \underline{3.20} & \underline{3.80} & 3.00 & \underline{2.67} & 1.83 & 2.88 & \underline{3.00} & \underline{2.50} & 2.45 & 2.73 & 2.73 \\
Seedance 1.5 Pro~\cite{seedance1p5} & 2.80 & 2.90 & 2.90 & \underline{3.13} & 2.50 & \underline{2.88} & \underline{3.10} & \underline{3.00} & 2.40 & \underline{2.85} & \underline{2.85} & \underline{3.23} \\
Kling 3.0~\cite{kling3.0}        & \underline{3.30} & 3.10 & 3.30 & 2.88 & 2.50 & \underline{2.88} & 2.90 & 2.80 & 2.30 & 2.69 & 2.62 & 3.15 \\
Seedance 2.0     & \textbf{3.90} & \textbf{3.60} & \textbf{4.10} & \textbf{3.75} & \textbf{3.75} & \textbf{3.88} & \textbf{3.30} & \textbf{3.20} & \textbf{3.30} & \textbf{3.54} & \textbf{3.54} & \textbf{3.54} \\
\bottomrule
\end{tabular}%
}
\end{table}

\paragraph{Voice Composite.} Seedance 2.0 leads on all four composite voice sub-categories in Table~\ref{tab:i2v_aud_composite}. Singing, rap, and instrumental audio across languages perform well, with melodies matched to the prompt context. Singing/rap scores 4.10 on APF---the model generates lyrics and melodies that match the prompted style. Off-screen voice scores 3.75/3.75/3.88, with audio-visual rhythm staying tight even when the speaker is not visible. Veo 3.1 is competitive on singing APF (3.80), its best result across all composite tasks, but drops to 1.83 on off-screen voice APF. Kling 3.0 scores 3.30 on singing AQ but only 2.30 on spatial scene APF. Audio-visual sync on off-screen narration is a pain point for most competitors---Kling 3.0 and Seedance 1.5 Pro both score 2.50 on AVS.

\begin{table}[ht!]
\centering
\caption{Fine-grained I2V audio evaluation on sound effects. Rating from 1 to 5.}
\vspace{-0.2cm}
\label{tab:i2v_aud_sfx}
\fittowidth{%
\begin{tabular}{l ccccccccccccccc}
\toprule
\multicolumn{1}{c}{\multirow{2}{*}{\rule{0pt}{3ex}\textbf{Model}}} & \multicolumn{3}{c}{\textbf{\makecell{Dialogue-\\Interaction}}} & \multicolumn{3}{c}{\textbf{\makecell{Object-Physical\\Events}}} & \multicolumn{3}{c}{\textbf{\makecell{Animal\\Sound}}} & \multicolumn{3}{c}{\textbf{\makecell{Background /\\Ambient}}} & \multicolumn{3}{c}{\textbf{\makecell{Special Effects\\(ASMR, etc.)}}} \\
\cmidrule(lr){2-4}\cmidrule(lr){5-7}\cmidrule(lr){8-10}\cmidrule(lr){11-13}\cmidrule(lr){14-16}
& \textbf{AQ} & \textbf{AVS} & \textbf{APF} & \textbf{AQ} & \textbf{AVS} & \textbf{APF} & \textbf{AQ} & \textbf{AVS} & \textbf{APF} & \textbf{AQ} & \textbf{AVS} & \textbf{APF} & \textbf{AQ} & \textbf{AVS} & \textbf{APF} \\
\midrule
Kling 2.6~\cite{kling2p6}        & 2.00 & 2.00 & 2.00 & 2.46 & 2.38 & 2.62 & 2.00 & 1.89 & 0.56 & 1.60 & 1.90 & 2.30 & 2.25 & 2.44 & 2.38 \\
Wan 2.6~\cite{wan2p6}          & 1.86 & 2.14 & 2.86 & 1.92 & 2.23 & 2.23 & 2.00 & 1.89 & 2.89 & 2.22 & 2.33 & 2.89 & 2.00 & 2.00 & 2.33 \\
Veo 3.1~\cite{veo3p1}          & 2.50 & 2.50 & 2.75 & 2.79 & 2.57 & \underline{3.14} & 2.33 & 2.44 & 2.78 & 2.50 & 2.50 & 3.20 & 2.86 & 2.71 & \underline{3.14} \\
Seedance 1.5 Pro~\cite{seedance1p5} & \underline{3.00} & \underline{2.75} & \underline{3.13} & \underline{2.86} & \underline{2.86} & 3.07 & \underline{3.11} & \underline{3.00} & \underline{3.11} & \textbf{3.30} & \underline{3.00} & \underline{3.60} & \underline{3.13} & \underline{3.19} & 2.94 \\
Kling 3.0~\cite{kling3.0}        & 2.88 & 2.25 & 3.00 & 2.77 & 2.69 & 3.00 & 2.78 & 2.78 & 3.00 & 2.80 & 2.90 & 3.30 & 2.75 & 2.75 & 2.63 \\
Seedance 2.0     & \textbf{3.50} & \textbf{3.00} & \textbf{3.88} & \textbf{3.57} & \textbf{3.50} & \textbf{3.86} & \textbf{3.56} & \textbf{3.22} & \textbf{3.44} & \underline{3.20} & \textbf{3.50} & \textbf{4.00} & \textbf{3.44} & \textbf{3.56} & \textbf{3.81} \\
\bottomrule
\end{tabular}%
}
\vspace{-0.2cm}
\end{table}

\paragraph{Sound Effects.} Seedance 2.0 leads on AVS for all five sound effects sub-categories and on AQ for four of five in Table~\ref{tab:i2v_aud_sfx}; Seedance 1.5 Pro edges ahead on background/ambient AQ (3.30 vs.\ 3.20). Voice, sound effects, and audio are well-layered---outputs sound like composed audio rather than isolated tracks stacked on top of each other. Background/ambient sound reaches 4.00 on APF---the model matches BGM and environmental audio to the video rhythm. Object-physical events scores 3.86 on APF, with action sounds synchronized to on-screen motion. Seedance 1.5 Pro is competitive on background sound (APF 3.60) and animal sound (AQ 3.11), placing it second on several sub-categories. Kling 2.6 scores 0.56 on animal sound APF---nearly zero prompt following---the lowest value across all audio tables. Wan 2.6 also struggles, scoring 2.00 on AQ for animal sound and dropping to 1.92 on object-physical events AQ.

\begin{table}[ht!]
\centering
\caption{Fine-grained I2V audio evaluation on instruments, dual-channel audio, and UGC creative tasks.}
\label{tab:i2v_aud_other}
\fittowidth{%
\begin{tabular}{l ccccccccc}
\toprule
\multicolumn{1}{c}{\multirow{2}{*}{\rule{0pt}{3ex}\textbf{Model}}} & \multicolumn{3}{c}{\textbf{\makecell{Instruments\\\ \& Audio}}} & \multicolumn{3}{c}{\textbf{\makecell{Dual-Channel\\Audio}}} & \multicolumn{3}{c}{\textbf{\makecell{UGC Creative /\\Portrait}}} \\
\cmidrule(lr){2-4}\cmidrule(lr){5-7}\cmidrule(lr){8-10}
& \textbf{AQ} & \textbf{AVS} & \textbf{APF} & \textbf{AQ} & \textbf{AVS} & \textbf{APF} & \textbf{AQ} & \textbf{AVS} & \textbf{APF} \\
\midrule
Kling 2.6~\cite{kling2p6}        & 2.50 & 2.86 & 2.57 & 2.00 & 2.07 & 2.07 & 2.27 & 2.36 & 2.45 \\
Wan 2.6~\cite{wan2p6}          & 2.29 & 2.36 & 2.64 & 2.08 & 1.92 & 2.38 & 2.09 & 2.45 & 2.73 \\
Veo 3.1~\cite{veo3p1}          & 2.87 & \underline{3.07} & \underline{3.00} & 2.43 & 2.71 & 2.36 & 2.64 & 2.55 & 3.00 \\
Seedance 1.5 Pro~\cite{seedance1p5} & \underline{3.13} & \underline{3.07} & 2.73 & \underline{3.13} & \underline{2.93} & \underline{2.87} & \underline{3.00} & 2.64 & \underline{3.45} \\
Kling 3.0~\cite{kling3.0}        & 3.00 & 2.67 & \underline{3.00} & 2.67 & 2.80 & 2.47 & 2.91 & \underline{2.91} & 3.00 \\
Seedance 2.0     & \textbf{3.60} & \textbf{3.27} & \textbf{3.80} & \textbf{3.47} & \textbf{3.53} & \textbf{3.27} & \textbf{3.64} & \textbf{3.64} & \textbf{3.82} \\
\bottomrule
\end{tabular}%
}
\end{table}

\paragraph{Other Audio.} In Table~\ref{tab:i2v_aud_other}, Seedance 2.0 scores 3.80 on instruments \& audio APF, generating instrument sounds and melodies that match the prompt. Dual-channel AVS reaches 3.53, with stereo separation that tracks the visual scene. UGC creative/portrait scores 3.64/3.64/3.82, its strongest sub-category in this table. Seedance 1.5 Pro is second on UGC APF (3.45) and instruments AQ (3.13). Dual-channel audio is weak across the board for competitors: Wan 2.6 scores 1.92 on AVS and Kling 2.6 scores 2.07, both below usability. The audio average row confirms the overall ranking: Seedance 2.0 (3.61/3.54/3.70), Seedance 1.5 Pro (3.07/2.95/3.10), Kling 3.0 (2.89/2.83/2.85), Veo 3.1 (2.68/2.69/2.79), Kling 2.6 (2.21/2.27/2.21), Wan 2.6 (2.20/2.18/2.55).

%% file: tables/r2v/r2v_score.tex
\begin{table}[ht!]
  \centering
  \caption{Reference-to-video (R2V) evaluation results. Multimodal Task Following and Prompt Following are rated 1--3, other dimensions are rated 1--5.}
  \vspace{-0.2cm}
  \label{tab:r2v_eval}
  \fittowidth{%
  \begin{tabular}{l ccccc}
  \toprule
  \multicolumn{1}{c}{\multirow{2}{*}{\textbf{Model}}} & \textbf{Multimodal Task} & \textbf{Editing} & \textbf{Reference} & \textbf{Motion} & \textbf{Prompt} \\
  & \textbf{Following} & \textbf{Consistency} & \textbf{Alignment} & \textbf{Quality} & \textbf{Following} \\
  \midrule
  Vidu Q2 Pro~\cite{viduq2}    & 2.13 & 2.29 & 1.79 & \underline{2.38} & \underline{2.08} \\
  Kling O1~\cite{klingO1}       & 2.30 & 2.89 & \underline{2.32} & 2.30 & 1.95 \\
  Kling 3.0~\cite{kling3.0}      & \underline{2.32} & \underline{3.37} & 2.37 & 2.36 & 1.95 \\
  Seedance 2.0   & \textbf{2.50} & \textbf{3.54} & \textbf{3.03} & \textbf{3.24} & \textbf{2.52} \\
  \bottomrule
  \end{tabular}%
  }
  \vspace{-0.3cm}
\end{table}

%% file: tables/r2v/r2v_support_matrix.tex
\begin{table}[H]
\centering
\caption{R2V multi-modal task support across models. \cmark\ = supported, \xmark\ = not supported.}
\vspace{-0.2cm}
\label{tab:r2v_task_support}
\resizebox{\textwidth}{!}{%
\begin{tabular}{llcccc}
\toprule
\textbf{Task} & \textbf{Input Modality} & \textbf{Seedance 2.0} & \textbf{Kling 3 Omni}~\cite{kling3.0} & \textbf{Vidu Q2 Pro}~\cite{viduq2} & \textbf{Kling O1}~\cite{klingO1} \\
\midrule
\multirow{5}{*}{\shortstack[l]{Subject\\Reference}}
& Image Reference                          & \cmark & \cmark & \cmark & \cmark \\
& Video Reference                          & \cmark & \cmark & \cmark & \xmark \\
& Audio-Visual Reference                   & \cmark & \cmark & \cmark & \xmark \\
& Audio + Image Reference                  & \cmark & \cmark & \cmark & \xmark \\
\midrule
\multirow{3}{*}{\shortstack[l]{Motion\\Reference}}
& Video Motion Reference                   & \cmark & \cmark & \cmark & \cmark \\
& Video Motion Reference + Image Reference & \cmark & \cmark & \cmark & \cmark \\
& Video Motion Reference + First Frame     & \cmark & \cmark & \cmark & \cmark \\
\midrule
\multirow{3}{*}{\shortstack[l]{Visual Effects /\\Creative Ref.}}
& Visual Effects / Creative Reference                    & \cmark & \xmark & \xmark & \xmark \\
& Visual Effects / Creative Reference + Image Reference  & \cmark & \xmark & \xmark & \xmark \\
& Visual Effects / Creative Reference + First Frame      & \cmark & \xmark & \xmark & \xmark \\
\midrule
\multirow{4}{*}{\shortstack[l]{Style\\Reference}}
& Style Image Reference                    & \cmark & \xmark & \cmark & \cmark \\
& Style Image + Subject Image Reference    & \cmark & \xmark & \cmark & \cmark \\
& Style Video Reference                    & \cmark & \xmark & \cmark & \cmark \\
& Style Video + Subject Image Reference    & \cmark & \xmark & \cmark & \cmark \\
\midrule
\multirow{3}{*}{\shortstack[l]{Video\\Editing}}
& Video Instruction Editing                & \cmark & \cmark & \cmark & \cmark \\
& Video Reference Image Editing            & \cmark & \cmark & \cmark & \cmark \\
\midrule
\multirow{4}{*}{\shortstack[l]{Continuation\\/ Extension}}
& Continuation                             & \cmark & \xmark & \xmark & \xmark \\
& Continuation + Subject Image Reference   & \cmark & \xmark & \xmark & \xmark \\
& Extension                                & \cmark & \xmark & \xmark & \xmark \\
& Extension + Subject Image Reference      & \cmark & \xmark & \xmark & \xmark \\
\bottomrule
\end{tabular}%
}
\vspace{-0.3cm}
\end{table}

%% file: tables/r2v/r2v_analysis.tex
\begin{table}[ht!]
  \centering
  \caption{R2V subject reference evaluation results. ``--'' denotes unsupported tasks. Task Fol. (Multimodal Task Following) is rated 1--3, Ref. Align. (Reference Alignment) is rated 1--5.}
  \vspace{-0.2cm}
  \label{tab:r2v_subject_ref}
  \fittowidth{%
  \begin{tabular}{l cc cc cc cc}
  \toprule
  \multicolumn{1}{c}{\multirow{3}{*}{\textbf{Model}}} & \multicolumn{2}{c}{\textbf{Subject Ref. Image}} & \multicolumn{2}{c}{\textbf{Subject Ref. Video}} & \multicolumn{2}{c}{\textbf{Subject Ref. First Video}} & \multicolumn{2}{c}{\textbf{Subject Ref. Image \& Audio}} \\
  \cmidrule(lr){2-3}\cmidrule(lr){4-5}\cmidrule(lr){6-7}\cmidrule(lr){8-9}
  & \textbf{Task Fol.} & \textbf{Ref. Align.} & \textbf{Task Fol.} & \textbf{Ref. Align.} & \textbf{Task Fol.} & \textbf{Ref. Align.} & \textbf{Task Fol.} & \textbf{Ref. Align.} \\
  \midrule
  Veo 3.1~\cite{veo3p1}        & -- & -- & -- & -- & -- & -- & -- & -- \\
  Sora 2~\cite{sora2}         & -- & -- & -- & -- & \textbf{3.00} & \textbf{3.27} & -- & -- \\
  Wan 2.6~\cite{wan2p6}        & -- & -- & -- & -- & 2.68 & 2.42 & -- & -- \\
  Kling O1~\cite{klingO1}       & \underline{2.71} & \underline{2.71} & -- & -- & -- & -- & -- & -- \\
  Vidu Q2 Pro~\cite{viduq2}    & 2.58 & 1.91 & \underline{2.50} & \underline{2.00} & -- & -- & -- & -- \\
  Kling 3 Omni~\cite{kling3.0}   & 2.50 & 2.55 & 2.67 & 2.50 & \underline{2.91} & \underline{2.82} & \underline{2.11} & \underline{2.05} \\
  Seedance 2.0   & \textbf{2.80} & \textbf{3.18} & \textbf{2.95} & \textbf{3.35} & 2.89 & \textbf{3.27} & \textbf{2.29} & \textbf{2.37} \\
  \bottomrule
  \end{tabular}%
  }
  \vspace{-0.2cm}
\end{table}

Seedance 2.0 achieves the best overall subject reference quality in both appearance and voice across all multi-modal competitors, with a clear lead in appearance consistency. As shown in Table~\ref{tab:r2v_subject_ref}, on image-based subject reference, Seedance 2.0 scores 2.80 on task following (1--3 scale) with 100\% 2-point rate and 80\% 3-point rate, ahead of Kling O1 (2.71, 73.68\%), Vidu Q2 Pro (2.58, 69.70\%), and Kling 3 Omni (2.50, 50\%). The reference alignment gap is wider: Seedance 2.0 scores 3.18 vs.\ Kling O1 at 2.71 and Vidu Q2 Pro at 1.91, a 1.27-point deficit for the latter. On video-based subject reference, Seedance 2.0 scores 2.95 on task following with 95\% of outputs reaching 3 points, and 3.35 on reference alignment---Vidu Q2 Pro trails at 2.00, a 1.35-point gap. For first-video reference, Sora 2 leads on task following (3.00, with 100\% 3-point rate), while Seedance 2.0 (2.89) and Kling 3 Omni (2.91) are close behind; on reference alignment, Seedance 2.0 and Sora 2 tie at 3.27. Image \& audio combined reference is supported only by Seedance 2.0 and Kling 3 Omni, with Seedance 2.0 scoring 2.29 vs.\ 2.11 on task following---the low absolute scores for both models indicate that joint image-audio conditioning remains a difficult problem.

\begin{table}[ht!]
\centering
\caption{R2V motion and style reference evaluation results. ``--'' denotes unsupported tasks. Task Fol. (Multimodal Task Following) is rated 1--3, Ref. Align. (Reference Alignment) and First Frame Pres. are rated 1--5.}
\vspace{-0.2cm}
\label{tab:r2v_motion_style_ref}
\fittowidth{%
\begin{tabular}{l ccc cc}
\toprule
\multicolumn{1}{c}{\multirow{3}{*}{\textbf{Model}}} & \multicolumn{3}{c}{\textbf{Motion Reference}} & \multicolumn{2}{c}{\textbf{Style Reference}} \\
\cmidrule(lr){2-4}\cmidrule(lr){5-6}
& \textbf{Task Fol.} & \textbf{Ref. Align.} & \textbf{First Frame Pres.} & \textbf{Task Fol.} & \textbf{Ref. Align.} \\
\midrule
Veo 3.1~\cite{veo3p1}        & -- & -- & -- & -- & -- \\
Sora 2~\cite{sora2}         & -- & -- & -- & -- & -- \\
Wan 2.6~\cite{wan2p6}        & -- & -- & -- & -- & -- \\
Kling O1~\cite{klingO1}       & 2.19 & 1.68 & \underline{3.46} & 1.96 & 1.84 \\
Vidu Q2 Pro~\cite{viduq2}    & 1.92 & 1.14 & 2.57 & \underline{2.15} & \underline{1.85} \\
Kling 3 Omni~\cite{kling3.0}   & \underline{2.20} & \underline{1.97} & \textbf{4.31} & -- & -- \\
Seedance 2.0   & \textbf{2.60} & \textbf{2.64} & 2.71 & \textbf{2.57} & \textbf{2.37} \\
\bottomrule
\end{tabular}%
}
\end{table}

Seedance 2.0 produces the best reference alignment for body motion, visual effects, and creative sequences, while competitors frequently fail to reproduce referenced effects or capture complete body movements. Table~\ref{tab:r2v_motion_style_ref} shows Seedance 2.0 scoring 2.60 on motion reference task following, ahead of Kling 3 Omni (2.20), Kling O1 (2.19), and Vidu Q2 Pro (1.92). On reference alignment, the gap widens: Seedance 2.0 scores 2.64, while all competitors fall below 2.0---Vidu Q2 Pro scores only 1.14, meaning most outputs bear little resemblance to the reference motion. One exception: on first-frame preservation, Kling 3 Omni scores 4.31, well above Seedance 2.0's 2.71. Kling 3 Omni tends to keep the first frame nearly unchanged but produces weaker subsequent motion, while Seedance 2.0 generates more dynamic video at the cost of lower first-frame fidelity. On style reference, Seedance 2.0 leads with 2.57 on task following (60\% 3-point rate) vs.\ Vidu Q2 Pro (2.15, 33.33\%) and Kling O1 (1.96, 10.71\%). Kling 3 Omni does not support style reference at all. Reference alignment follows the same order: 2.37, 1.85, 1.84. When combining style and subject reference, Seedance 2.0 produces more accurate responses and better generation quality; competitors frequently misinterpret the task as reference-image editing or produce artifacts, with Kling 3 Omni exhibiting this issue most often.

\begin{table}[ht!]
\centering
\caption{R2V video editing, continuation, and extension evaluation results. ``--'' denotes unsupported tasks. Task Fol. (Multimodal Task Following) is rated 1--3, Ref. Align. (Reference Alignment), Edit. Consist. (Editing Consistency), and other dimensions are rated 1--5.}
\vspace{-0.2cm}
\label{tab:r2v_editing_continuation}
\fittowidth{%
\begin{tabular}{l ccc cc cc}
\toprule
\multicolumn{1}{c}{\multirow{3}{*}{\textbf{Model}}} & \multicolumn{3}{c}{\textbf{Video Editing}} & \multicolumn{2}{c}{\textbf{Video Continuation}} & \multicolumn{2}{c}{\textbf{Video Extension}} \\
\cmidrule(lr){2-4}\cmidrule(lr){5-6}\cmidrule(lr){7-8}
& \textbf{Task Fol.} & \textbf{Ref. Align.} & \textbf{Edit. Consist.} & \textbf{Task Fol.} & \textbf{Ref. Align.} & \textbf{Task Fol.} & \textbf{Ref. Align.} \\
\midrule
Veo 3.1~\cite{veo3p1}        & -- & -- & -- & -- & -- & \textbf{2.78} & \textbf{3.44} \\
Sora 2~\cite{sora2}         & -- & -- & -- & -- & -- & -- & -- \\
Wan 2.6~\cite{wan2p6}        & -- & -- & -- & -- & -- & -- & -- \\
Kling O1~\cite{klingO1}       & \textbf{2.29} & \underline{3.03} & \underline{2.78} & -- & -- & -- & -- \\
Vidu Q2 Pro~\cite{viduq2}    & 2.02 & 2.58 & 2.22 & -- & -- & -- & -- \\
Kling 3 Omni~\cite{kling3.0}   & \underline{2.24} & 2.71 & \underline{3.09} & -- & -- & -- & -- \\
Seedance 2.0   & 2.20 & \textbf{3.79} & \textbf{3.75} & \textbf{2.88} & \textbf{3.18} & \underline{1.93} & \underline{3.28} \\
\bottomrule
\end{tabular}%
}
\vspace{-0.3cm}
\end{table}

Video editing is the most competitive R2V task. In Table~\ref{tab:r2v_editing_continuation}, Kling O1 slightly leads on task following (2.29 vs.\ Seedance 2.0's 2.20), and Kling 3 Omni is close at 2.24---all three within 0.09 points. However, Seedance 2.0 pulls ahead on reference alignment (3.79 vs.\ Kling O1's 3.03) and editing consistency (3.75 vs.\ Kling 3 Omni's 3.09). Seedance 2.0 responds better to long-text and multi-edit instructions and handles complex editing tasks more completely, while also producing more accurate results for well-known IP references. All models share common failure modes: unresponsive edits and unintended modifications to non-edit regions.

Video continuation is currently supported only by Seedance 2.0, which scores 2.88 on task following and 3.18 on reference alignment. It handles complex narratives and long-text continuation prompts well, though issues remain with color consistency, multi-subject omission, and subject duplication.

For video extension, Seedance 2.0 and Veo 3.1 are the only two models evaluated, but they differ in scope: Seedance 2.0 accepts arbitrary uploaded videos for extension and supports combining extension with subject image input, while Veo 3.1 can only extend videos it generated itself. Despite broader input support, Seedance 2.0's extension quality trails Veo 3.1 notably---Veo 3.1 scores 2.78 on task following (88.89\% 3-point rate) vs.\ Seedance 2.0's 1.93 (31.82\%), and 3.44 vs.\ 3.28 on reference alignment, making extension Seedance 2.0's weakest R2V task.

%% file: sections/contributions.tex
\section{Contributions and Acknowledgments}
\label{contributions}

All authors of Seedance-2.0 are listed in alphabetical order by their last names.

\begin{multicols}{3} %
\sffamily{\color{seedblue}  \large{Team Seedance}} \\
\\
\\
\color{seedblue}De Chen\\
\color{seedblue}Liyang Chen\\
\color{seedblue}Xin Chen\\
\color{seedblue}Ying Chen\\
\color{seedblue}Zhuo Chen\\
\color{seedblue}Zhuowei Chen\\
\color{seedblue}Feng Cheng\\
\color{seedblue}Tianheng Cheng\\
\color{seedblue}Yufeng Cheng\\
\color{seedblue}Mojie Chi\\
\color{seedblue}Xuyan Chi\\
\color{seedblue}Jian Cong\\
\color{seedblue}Qinpeng Cui\\
\color{seedblue}Fei Ding\\
\color{seedblue}Qide Dong\\
\color{seedblue}Yujiao Du\\
\color{seedblue}Haojie Duanmu\\
\color{seedblue}Junliang Fan\\
\color{seedblue}Jiarui Fang\\
\color{seedblue}Jing Fang\\
\color{seedblue}Zetao Fang\\
\color{seedblue}Chengjian Feng\\
\color{seedblue}Yu Gao\\
\color{seedblue}Diandian Gu\\
\color{seedblue}Dong Guo\\
\color{seedblue}Hanzhong Guo\\
\color{seedblue}Qiushan Guo\\
\color{seedblue}Boyang Hao\\
\color{seedblue}Hongxiang Hao\\
\color{seedblue}Haoxun He\\
\color{seedblue}Jiaao He\\
\color{seedblue}Qian He\\
\color{seedblue}Tuyen Hoang\\
\color{seedblue}Heng Hu\\
\color{seedblue}Ruoqing Hu\\
\color{seedblue}Yuxiang Hu\\
\color{seedblue}Jiancheng Huang\\
\color{seedblue}Weilin Huang\\
\color{seedblue}Zhaoyang Huang\\
\color{seedblue}Zhongyi Huang\\
\color{seedblue}Jishuo Jin\\
\color{seedblue}Ming Jing\\
\color{seedblue}Ashley Kim\\
\color{seedblue}Shanshan Lao\\
\color{seedblue}Yichong Leng\\
\color{seedblue}Bingchuan Li\\
\color{seedblue}Gen Li\\
\color{seedblue}Haifeng Li\\
\color{seedblue}Huixia Li\\
\color{seedblue}Jiashi Li\\
\color{seedblue}Ming Li\\
\color{seedblue}Xiaojie Li\\
\color{seedblue}Xingxing Li\\
\color{seedblue}Yameng Li\\
\color{seedblue}Yiying Li\\
\color{seedblue}Yu Li\\
\color{seedblue}Yueyan Li\\
\color{seedblue}Chao Liang\\
\color{seedblue}Han Liang\\
\color{seedblue}Jianzhong Liang\\
\color{seedblue}Ying Liang\\
\color{seedblue}Wang Liao\\
\color{seedblue}J. H. Lien\\
\color{seedblue}Shanchuan Lin\\
\color{seedblue}Xi Lin\\
\color{seedblue}Feng Ling\\
\color{seedblue}Yue Ling\\
\color{seedblue}Fangfang Liu\\
\color{seedblue}Jiawei Liu\\
\color{seedblue}Jihao Liu\\
\color{seedblue}Jingtuo Liu\\
\color{seedblue}Shu Liu\\
\color{seedblue}Sichao Liu\\
\color{seedblue}Wei Liu\\
\color{seedblue}Xue Liu\\
\color{seedblue}Zuxi Liu\\
\color{seedblue}Ruijie Lu\\
\color{seedblue}Lecheng Lyu\\
\color{seedblue}Jingting Ma\\
\color{seedblue}Tianxiang Ma\\
\color{seedblue}Xiaonan Nie\\
\color{seedblue}Jingzhe Ning\\
\color{seedblue}Junjie Pan\\
\color{seedblue}Xitong Pan\\
\color{seedblue}Ronggui Peng\\
\color{seedblue}Xueqiong Qu\\
\color{seedblue}Yuxi Ren\\
\color{seedblue}Yuchen Shen\\
\color{seedblue}Guang Shi\\
\color{seedblue}Lei Shi\\
\color{seedblue}Yinglong Song\\
\color{seedblue}Fan Sun\\
\color{seedblue}Li Sun\\
\color{seedblue}Renfei Sun\\
\color{seedblue}Wenjing Tang\\
\color{seedblue}Boyang Tao\\
\color{seedblue}Zirui Tao\\
\color{seedblue}Dongliang Wang\\
\color{seedblue}Feng Wang\\
\color{seedblue}Hulin Wang\\
\color{seedblue}Ke Wang\\
\color{seedblue}Qingyi Wang\\
\color{seedblue}Rui Wang\\
\color{seedblue}Shuai Wang\\
\color{seedblue}Shulei Wang\\
\color{seedblue}Weichen Wang\\
\color{seedblue}Xuanda Wang\\
\color{seedblue}Yanhui Wang\\
\color{seedblue}Yue Wang\\
\color{seedblue}Yuping Wang\\
\color{seedblue}Yuxuan Wang\\
\color{seedblue}Zijie Wang\\
\color{seedblue}Ziyu Wang\\
\color{seedblue}Guoqiang Wei\\
\color{seedblue}Meng Wei\\
\color{seedblue}Di Wu\\
\color{seedblue}Guohong Wu\\
\color{seedblue}Hanjie Wu\\
\color{seedblue}Huachao Wu\\
\color{seedblue}Jian Wu\\
\color{seedblue}Jie Wu\\
\color{seedblue}Ruolan Wu\\
\color{seedblue}Shaojin Wu\\
\color{seedblue}Xiaohu Wu\\
\color{seedblue}Xinglong Wu\\
\color{seedblue}Yonghui Wu\\
\color{seedblue}Ruiqi Xia\\
\color{seedblue}Xin Xia\\
\color{seedblue}Xuefeng Xiao\\
\color{seedblue}Shuang Xu\\
\color{seedblue}Bangbang Yang\\
\color{seedblue}Jiaqi Yang\\
\color{seedblue}Runkai Yang\\
\color{seedblue}Tao Yang\\
\color{seedblue}Yihang Yang\\
\color{seedblue}Zhixian Yang\\
\color{seedblue}Ziyan Yang\\
\color{seedblue}Fulong Ye\\
\color{seedblue}Bingqian Yi\\
\color{seedblue}Xing Yin\\
\color{seedblue}Yongbin You\\
\color{seedblue}Linxiao Yuan\\
\color{seedblue}Weihong Zeng\\
\color{seedblue}Xuejiao Zeng\\
\color{seedblue}Yan Zeng\\
\color{seedblue}Siyu Zhai\\
\color{seedblue}Zhonghua Zhai\\
\color{seedblue}Bowen Zhang\\
\color{seedblue}Chenlin Zhang\\
\color{seedblue}Heng Zhang\\
\color{seedblue}Jun Zhang\\
\color{seedblue}Manlin Zhang\\
\color{seedblue}Peiyuan Zhang\\
\color{seedblue}Shuo Zhang\\
\color{seedblue}Xiaohe Zhang\\
\color{seedblue}Xiaoying Zhang\\
\color{seedblue}Xinyan Zhang\\
\color{seedblue}Xinyi Zhang\\
\color{seedblue}Yichi Zhang\\
\color{seedblue}Zixiang Zhang\\
\color{seedblue}Haiyu Zhao\\
\color{seedblue}Huating Zhao\\
\color{seedblue}Liming Zhao\\
\color{seedblue}Yian Zhao\\
\color{seedblue}Guangcong Zheng\\
\color{seedblue}Jianbin Zheng\\
\color{seedblue}Xiaozheng Zheng\\
\color{seedblue}Zerong Zheng\\
\color{seedblue}Kuan Zhu\\
\color{seedblue}Feilong Zuo\\

\end{multicols}